\documentclass[review]{elsarticle}
\usepackage{hyperref}

\usepackage{lineno}
\modulolinenumbers[5]

\journal{Journal of Visual Communication and Image Representation}
\usepackage{amssymb}

\usepackage{url,graphicx,amsmath,times,amssymb}
\usepackage{floatflt}

\newcommand{\var}{\text{var}}
\graphicspath{ {../figs/} {./}}

\begin{document}

\begin{frontmatter}

\title{RENOIR - A Dataset for Real Low-Light Image Noise Reduction}

\author[mymainaddress]{Josue Anaya}
\author[mymainaddress]{Adrian Barbu\corref{mycorrespondingauthor}}
\address{}

\ead[url]{http://ani.stat.fsu.edu/~abarbu/}

\cortext[mycorrespondingauthor]{Corresponding author.}
\ead{abarbu@stat.fsu.edu}

\address[mymainaddress]{Department of Statistics, Florida State University, 117 N Woodward Ave, Tallahassee FL 32306, USA}


\begin{abstract}

Image denoising algorithms are evaluated using images corrupted by artificial noise, which may lead to incorrect conclusions about their performances on real noise. In this paper we introduce a dataset of color images corrupted by natural noise due to low-light conditions, together with spatially and intensity-aligned low noise images of the same scenes. We also introduce a method for estimating the true noise level in our images, since even the low noise images contain small amounts of noise. We evaluate the accuracy of our noise estimation method on real and artificial noise, and investigate the Poisson-Gaussian noise model. Finally, we use our dataset to evaluate six denoising algorithms: Active Random Field, BM3D, Bilevel-MRF, Multi-Layer Perceptron, and two versions of NL-means. We show that while the Multi-Layer Perceptron, Bilevel-MRF, and NL-means with soft threshold  outperform BM3D on gray images with synthetic noise, they lag behind on our dataset.

\end{abstract}
\begin{keyword}
image denoising \sep denoising dataset \sep low light noise \sep Poisson-Gaussian noise model
\end{keyword}

\end{frontmatter}


\section{Introduction and Motivation}

In the field of computer vision and computational photography, noise reduction is the application in which granular discrepancies found in images are removed. The task of performing noise reduction is synonymous with improvement in image quality. Many consumer  cameras and mobile phones deal with the issues of low-light noise due to small sensor size and insufficient exposure time. The issue of noise for a particular digital camera is so important that it is used as a valuable metric of the camera sensor and for comparing camera performance\cite{dxomark}. 

Besides the noise in digital camera images, another example of images that deal with noise due to limited acquisition time are Magnetic Resonance Images (MRI). Other important types of image modalities such as X-ray and CT (Computed Tomography) also suffer from noise artifacts due to insufficient exposure because of low radiation dose limits.  While the image acquisition process is different in all of these examples, the reason for the noise is in most part the same. This is why the problem of low-light image noise reduction is studied and has led to a variety of different noise reduction methods \cite{buades2005non,portilla2003image,dabov2007image,barbu2009training,StochasticImageDenoising,mairal2009non,NeuralNetworkCompete,schmidt2010generative,optMRF}.

In general most of the performance evaluations for these various noise reduction methods are done on small images contaminated with noise of a known type (Gaussian, Poisson, salt and pepper, etc), which is artificially added to a clean image to obtain a noisy version. Measuring the performance of a noise reduction algorithm on small images corrupted by artificial noise might not give an accurate enough picture of the denoising performance of the algorithm on real digital camera images in low-light conditions. The nature of the noise in low-light camera images is more complex than just i.i.d., for example its variance depends on the image intensity \cite{Foi-Poisson} and has small-range correlations \cite{NoiseEstimation}, so it would be desirable to obtain images naturally corrupted by low-light noise and their noise-free counterparts. For this purpose, we bring the following contributions:

\begin{itemize}
\item A dataset of color images naturally corrupted by low-light noise, taken with two digital cameras (Canon PowerShot S90, Canon EOS Rebel T3i) and a mobile phone camera (Xiaomi Mi3).
\item A process for the collection of noisy and low-noise pixel-aligned  images of the same scene.
\item A method for aligning the intensity values of all the images of the same scene.
\item A technique for computing the noise level and PSNR (Peak Signal-to-Noise Ratio) \cite{perceptualdistortion} of the images in our dataset and an evaluation of its accuracy.
\item An evaluation of the Poisson-Gaussian mixture model \cite{Foi-Poisson,makitalo2014noise} and its parameter estimation method.
\item An evaluation of the denoising performance of six algorithms on our dataset, with some surprising results.
\end{itemize}

Different cameras produce different kinds of noise due to their sensor size, sensor type, and other factors on the imaging pipeline. A learning-based denoising method (e.g. \cite{barbu2009training, NeuralNetworkCompete} or \cite{optMRF}) could be trained for a specific type of camera on noisy-clean image pairs for that specific camera, but it is not clear how well it would generalize to images from another camera. Trying to construct a dataset of various image pairs from  different cameras may help determine which denoising method generalizes well over many different cameras, however it does not evaluate the full potential of a method on any one specific camera at various noise levels. We therefore selected to obtain an equal number of images from three cameras with different sensor sizes: one with a small sensor (Xiaomi Mi3), one with a slightly larger sensor  (Canon S90) and one with a mid-size sensor (Canon T3i) and obtain many images with different noise levels for each camera.

\subsection{Related Works}
 The Tampere Image Database (TID 2013) \cite{ponomarenko2015image} is a dataset intended for evaluating image quality assessment metrics such as the SSIM \cite{ssim}. It contains 25 reference images and 3000 images obtained by corrupting the reference images by 24 types of noise and distortions such additive or multiplicative Gaussian noise, high frequency noise, image encoding artifacts, image denoising artifacts, etc. However, these corrupted images are all obtained by artificially transforming the reference images in different ways. In contrast, our dataset provides images where the noise is obtained naturally by short time exposure to low-light scenes and clean images obtained by long time exposure to the same scene. Besides evaluating quality assessment metrics, our dataset could be used for other applications such as studying noise formation and noise statistics in digital camera images, and for training or evaluating image denoising algorithms.
 
The only database for benchmarking image denoising that we are aware of  is \cite{ImageDenoisingBenchmark} (discussed in \cite{StochasticImageDenoising}). It evaluates various denoising methods on color images corrupted by artificial Gaussian noise. The problem with artificial Gaussian noise is  that it is a very simple noise model that is not present in real world images where many times the noise level changes with the image intensity. In this respect it has been shown that the distribution of low-light camera noise is not Gaussian, but follows a more complex Poisson-Gaussian mixture distribution with intensity dependent variance \cite{Foi-Poisson,Luisier2010d}. Our dataset contains images corrupted by real low-light noise, which besides offering a more realistic setting for image denoising, it allows to see how well the Poisson-Gaussian model fits real noisy images obtained in low-light conditions. We will show such a study in Section \ref{sec:poisson}.
 
A small number of real low-light noise images with corresponding clean image pairs have been used in \cite{NoiseEstimation} to study the noise and intensity relationship. 
However, we are not aware of any public database or collection of images that have been corrupted by real low-light noise like the ones presented in our paper, and which took all the necessary steps for carefully acquiring the images, intensity aligning them and diagnosing the quality of the obtained pairs.

A dataset with real low-light noisy images and their clean counterparts such as the one introduced in this paper would bring many benefits to just using images artificially corrupted by noise from a Poisson-Gaussian mixture model:
\begin{itemize}
\item It would contribute to the further study of the noise structure in digital cameras, such as how much it differs from the Poisson-Gaussian mixture model, spatial correlation structure, how noise parameters relate to camera acquisition parameters, etc.
\item It would present a more realistic range of levels of noise, similar to what happens ``in the wild''. In contrast, the Poisson-Gaussian noise model is usually studied with fixed (and known) noise parameters, such as in \cite{makitalo2014noise}. 
\item It would allow for an end-to-end evaluation of denoising algorithms. Evaluating using artificial noise models, even if they are accurate, might not entirely reflect the reality of noise in digital cameras.
\end{itemize}

\section{Acquisition of Natural Image Pairs} \label{sec:acq}

The dataset \footnote{Available at: \url{http://adrianbarburesearch.blogspot.com/p/renoir-dataset.html}} acquired in this paper consists low-light uncompressed natural images of 120 scenes. About four images per scene were acquired, where two images contain noise and the other two images contain very little noise. The presence of noise in  the images is mainly due to the number of photons that are received by the camera's sensor and the amplification process, as discussed in \cite{ishii2007denoising}. 

\subsection{Acquisition procedure}

 All the images in our dataset are of static scenes and are acquired under low-light  conditions using the following "sandwich" procedure:
\begin{itemize}
\item A low-noise image is obtained with low light sensitivity (ISO 100) and long exposure time. This will be the {\em reference} image.
\item One or two noisy images are then obtained with increased light sensitivity and reduced exposure time. 
\item Finally, another low noise image is taken with the same parameters as the reference image.  This will be the {\em clean} image.
\end{itemize}

The two low-noise (reference and clean) images are acquired at the beginning and at the end of the sequence, while the one or two noisy images are shot in between. This is done to evaluate the quality of the whole acquisition process for that particular scene. This process is somehow similar to the process discussed in \cite{NoiseEstimation} which used pair images that were taken with flash. The problem with taking the images with flash is that the flash can change the scene illumination. Moreover, in  \cite{NoiseEstimation} no brightness alignment has been performed on their images. 
The two low-noise images are used as a first level of quality control. Any motion or lighting change during acquisition could make the two low-noise images be sufficiently different, as measured by the PSNR. In fact, we discarded the scenes with PSNR of the clean images less than 34.

The actual acquisition parameters for each camera are presented in Table \ref{tab:camerainfo}.
\begin{table}[htb]
\vspace{-4mm}
\caption{ISO (and Exposure time) per camera \label{tab:camerainfo}}
\vskip -0mm
\centering
\begin{tabular}{|l |c c|c c|}
\hline	
&\multicolumn{2}{c}{ Reference/Clean Images }	&\multicolumn{2}{|c|}{Noisy Images}\\
Camera & ISO &Time(s) & ISO &Time(s)\\
\hline
Mi3\phantom{$I^I$}		&100  &auto	&1600 or 3200  &auto	\\
S90		&100 &3.2	&640 or 1000 &auto	\\
T3i 		&100 &auto	&3200 or 6400 &auto	\\
\hline
\end{tabular}
\vspace{-4mm}
\end{table}

Using the Canon Developer Tool Kit for the Canon S90 and the EOS Utility for the Canon Rebel T3i we were able to program the automatic collection of the four images while trying to preserve the static scene in the images by not moving or refocusing the camera.  The sandwich approach that we used to obtain our images also helped insure that the only visual difference between the images of a scene was simply due to noise.  All of the images were collected and saved in RAW format (CR2 for Canon). The Mi2Raw Camera app was used to capture the RAW images for the Xiaomi Mi3 (in DNG format).
\begin{figure*}[htb]
\centering
\includegraphics[height=4.5cm]{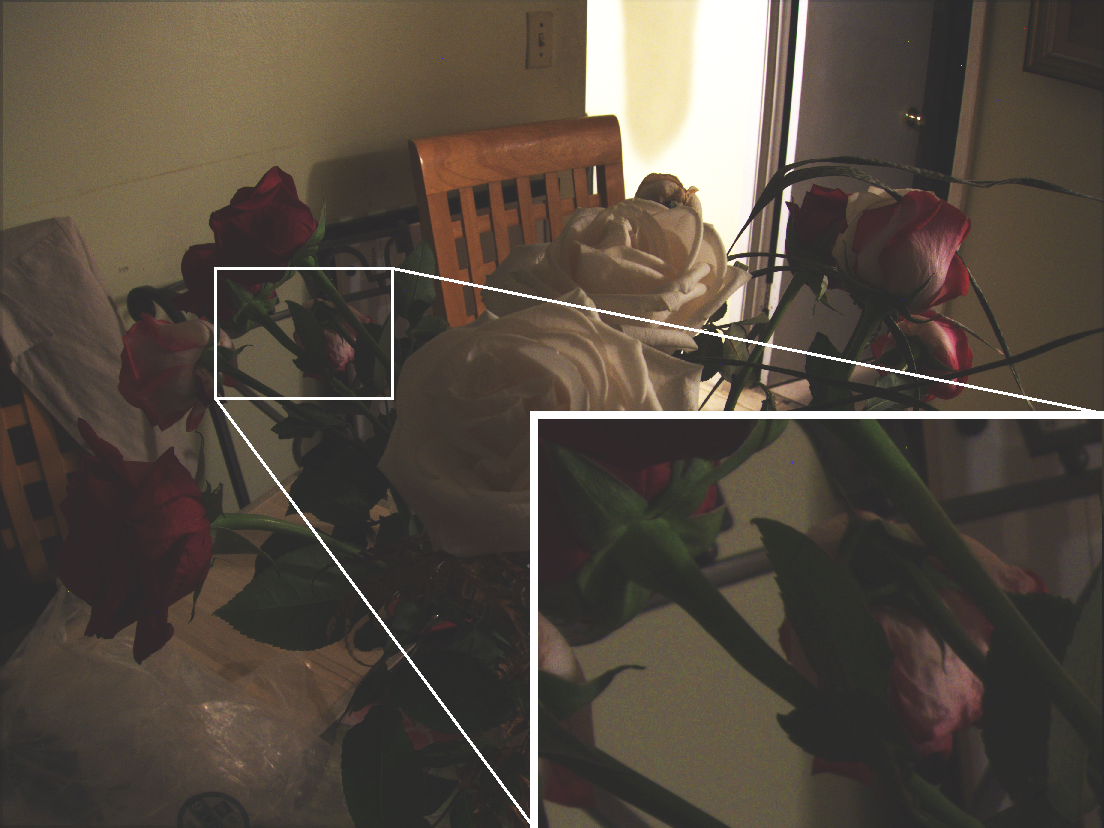}
\includegraphics[height=4.5cm]{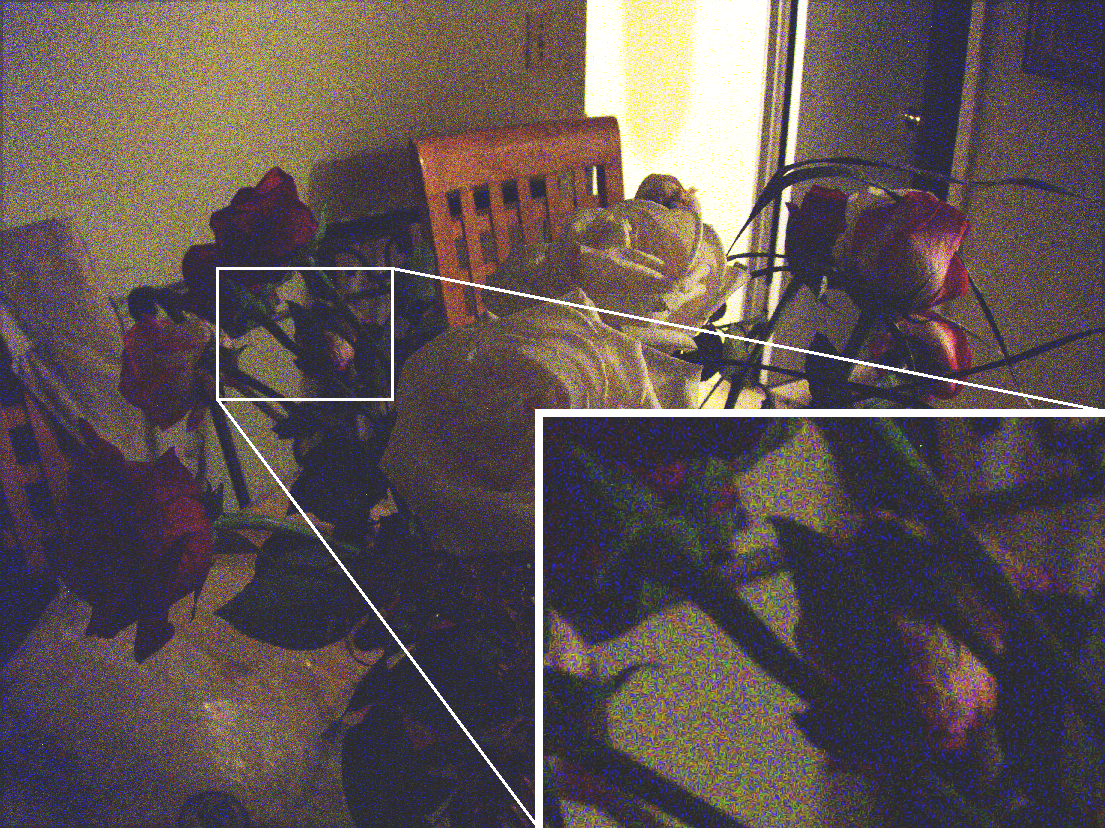}
\vskip 0.5mm
\includegraphics[height=4.5cm]{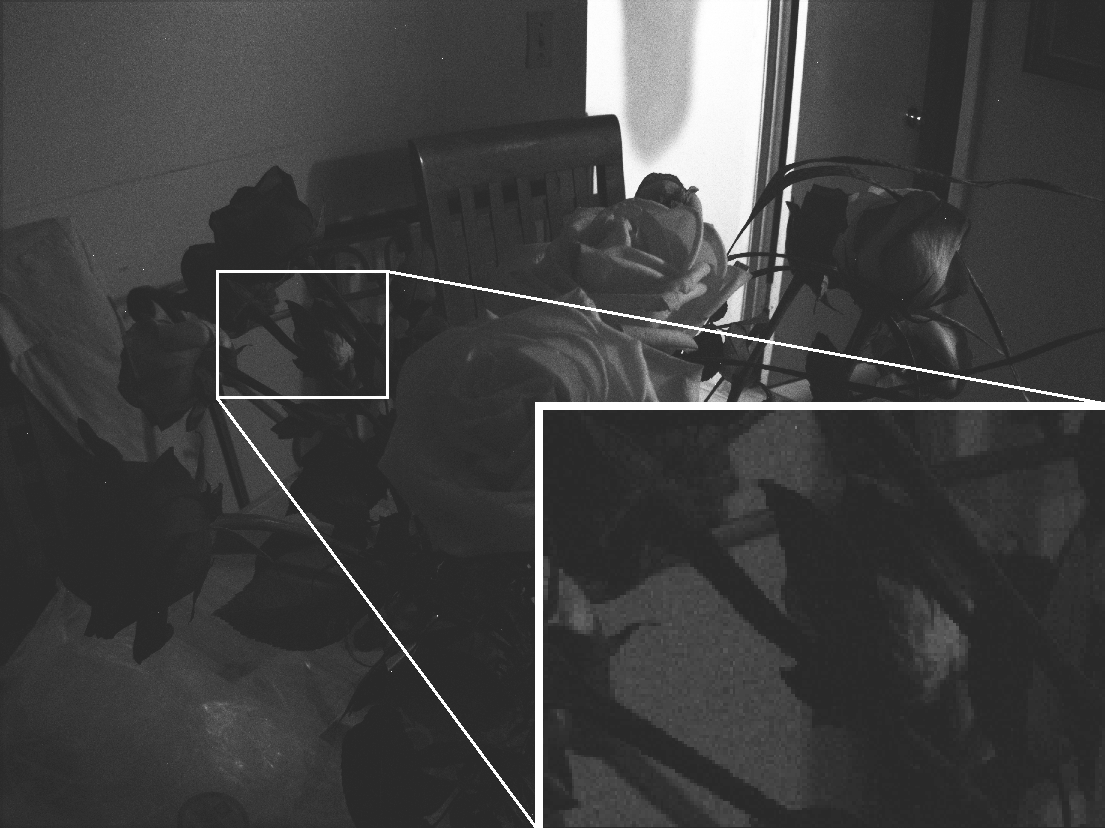}
\includegraphics[height=4.5cm]{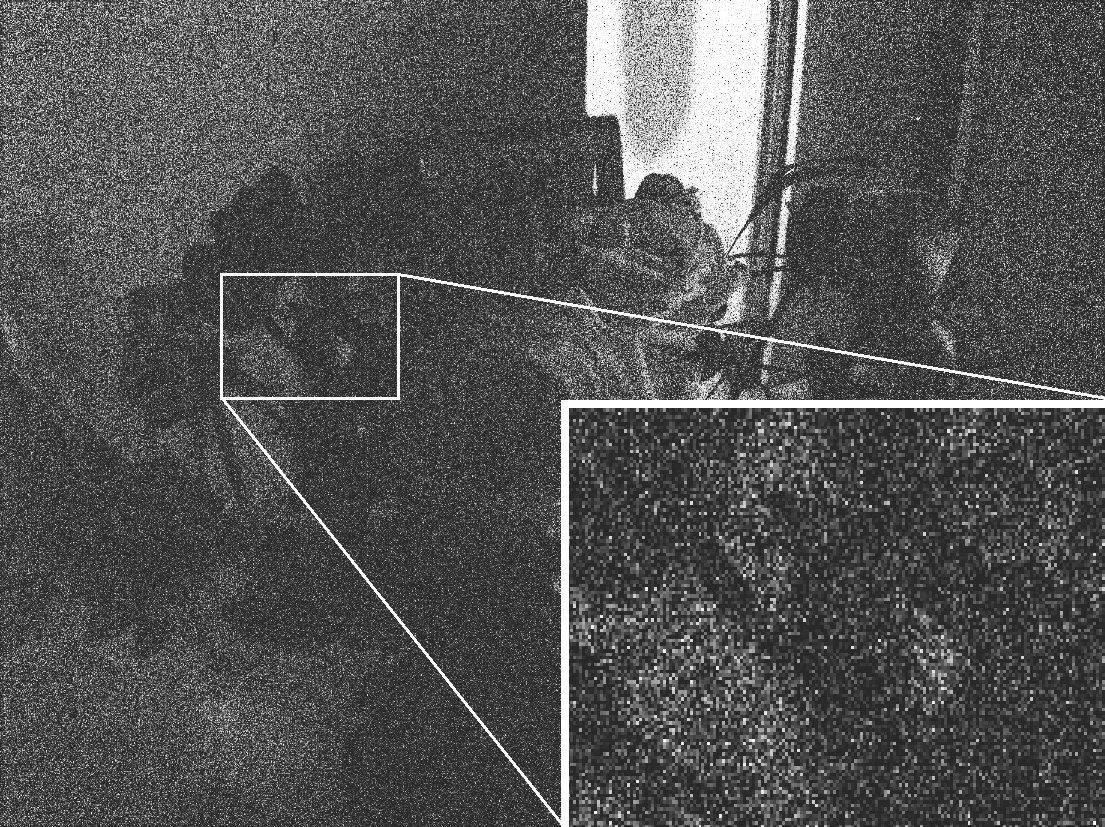}
\vskip -3mm
\caption{An example of a clean and noisy image pair as well as their corresponding blue channel. The noise present is the result of the low-light environment. The images were taken using a Canon PowerShot S90. }
\label {fig:s90example}
\vspace{-2mm}
\end{figure*}

An example of one of the images in the dataset can be seen in Figure \ref{fig:s90example}. In the end we collected 40 scenes for the S90, 40 for the T3i, and another 40 for the Mi3.

The image denoising database in \cite{ImageDenoisingBenchmark} contains 300 noisy images at 5 noise levels ($ \sigma =5,10,15,25,35$) for a total of 1500 images. The dimensionality of these images is 481 by 321 . The dimensionality of the images for just the S90 images is 3684 by 2760 while the images from the other cameras are even larger, as shown in Table \ref{tab:datastats}. Although our image database contains fewer noisy images, our images contain about 60 times more pixels and therefore more patch variability for studying noise models from just one of the three cameras.

Many various denoising methods \cite{barbu2009training,NeuralNetworkCompete,optMRF} train models from noisy-clean image pairs that are supposed to generalize well to future noisy images. For this reason and for evaluation in general it is very important to maintain a careful construction of these noisy-clean image pairs and to have many examples for a representative dataset. The difficulty in constructing such pairs is why artificial noise is used in practice.

\subsection{Mobile camera difficulties}

\begin{figure*}[htb]
\vspace{-3mm}
\centering
\hspace{-2mm}
\includegraphics[width=4.cm, height=4.cm]{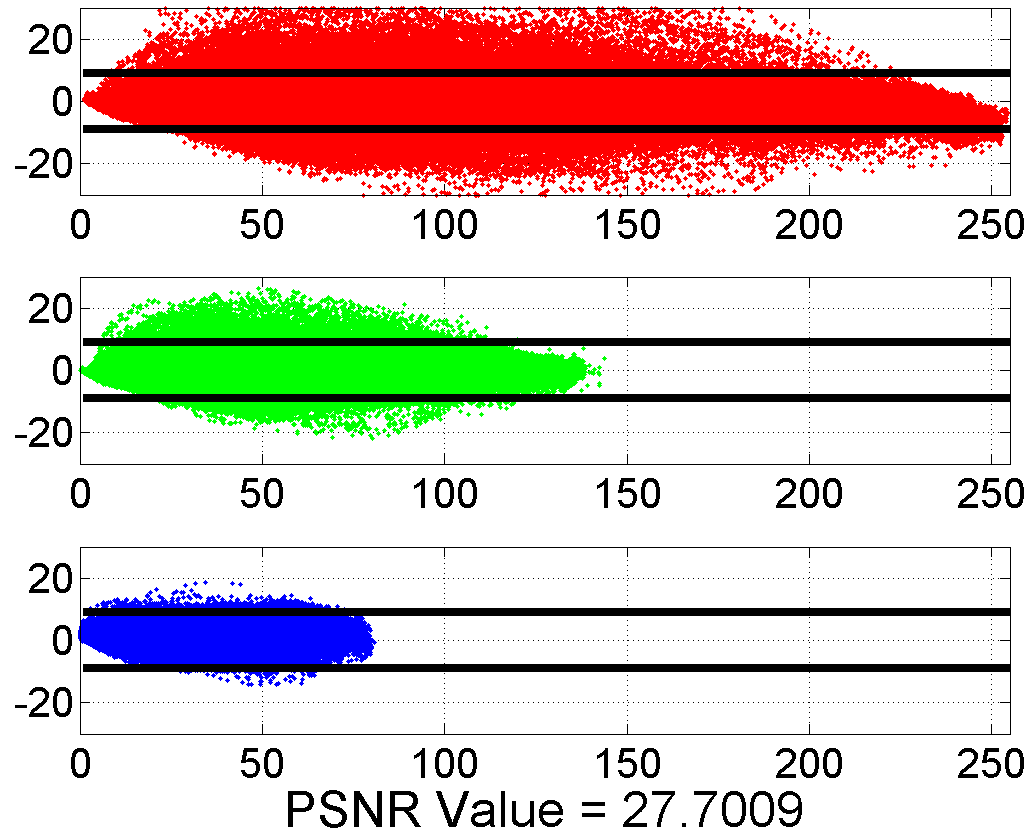}
\hspace{-2mm}
\includegraphics[width=4.cm, height=4.cm]{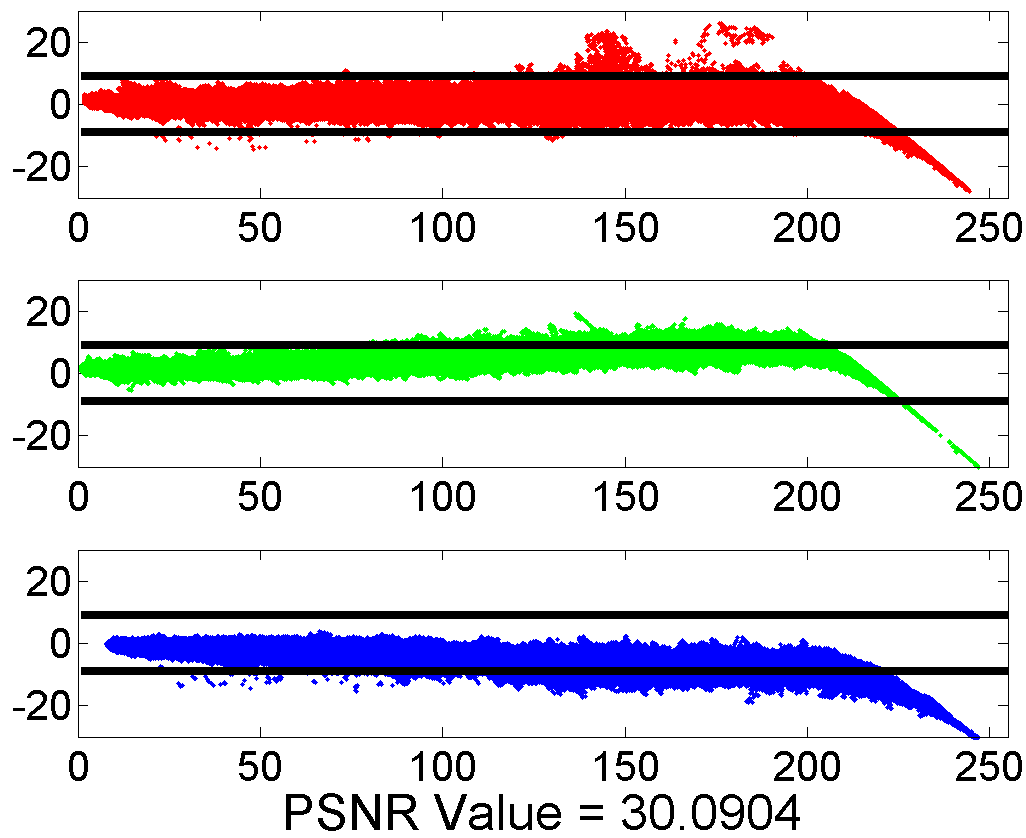}
\hspace{-2mm}
\includegraphics[width=4.cm, height=4.cm]{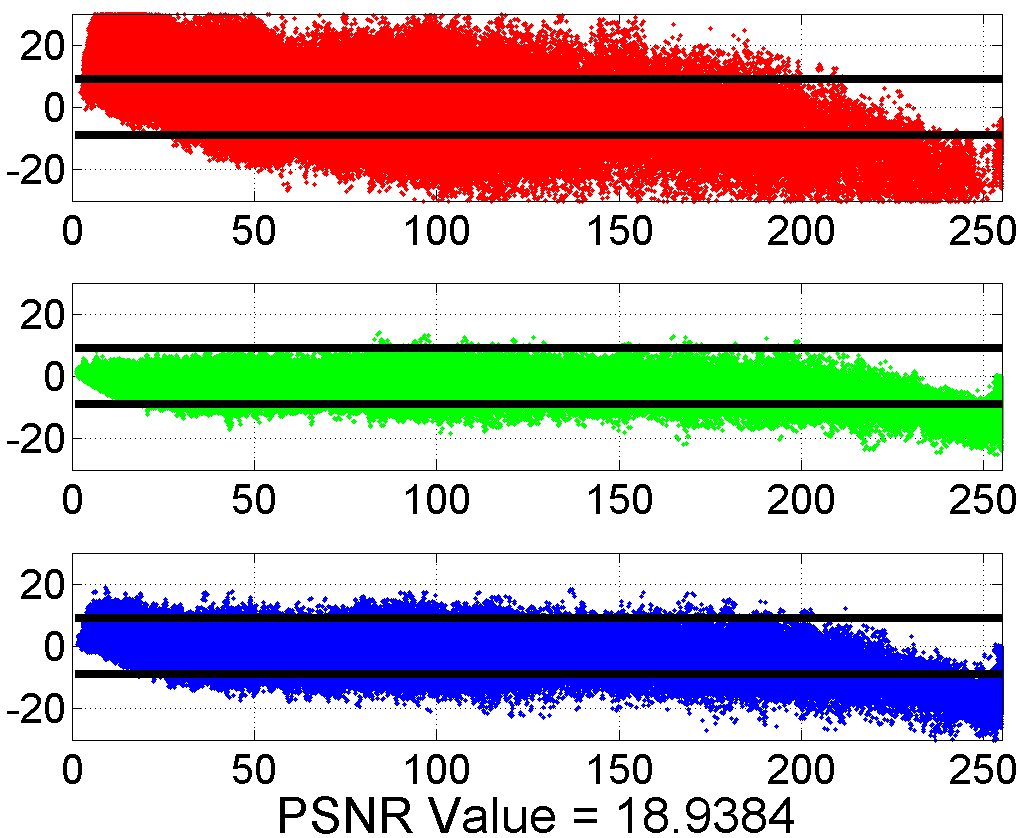}
\vskip-5mm
\caption{Intensity alignment issues observed on scatter plots of the intensity difference between the reference image and aligned noisy image vs reference image intensity. Left: image movement during the 'sandwich' procedure. Middle: light saturation. Right: non-linearity at high ISO levels.}\label {fig:Mi3issues}
\vspace{-4mm}
\end{figure*}

In trying to collect images for our dataset we decided on also collecting mobile phone camera images. In doing so we ran into many of difficulties, some of which are described below and are illustrated in Figure \ref{fig:Mi3issues}. 

The first difficulty that arose was collecting the RAW images on phone cameras. Only a few mobile phones collect true RAW images (data dump directly from the sensor). For example, the Iphone can have an application installed that will allow it to take RAW images, however these RAW images are not in fact truly RAW because the sensor data has gone through some unknown post processing. Therefore, only some of the most recent phones that have been allowed by the device manufacturer can truly collect RAW images.

The second difficulty that we found when trying to collect mobile phone images came in the control over certain image acquisition parameters such as the exposure time and ISO values. These mobile phone cameras already have a very small sensor, so when we tried to use a Google Nexus 5 with the  FV-5 camera application to capture RAW images, the limits of control over settings like the exposure time and ISO, and its tiny sensor size led to many scenes failing our 'sandwich' procedure selection benchmark ( did not have sufficient amount of light needed for the PSNR to be around 35 for the reference and clean image.) We also noted for a phone like the Google Nexus 5 a non-linearity relationship issue in the brightness alignment procedure (this could also be due to an insufficient amount of light.) 

\begin{table*}[htb]
\vspace{-3mm}
\centering
\caption {Description of the dataset and size \label{tab:datastats}}
\vskip -1mm
\scalebox{0.95}{
\begin{tabular}{|l*{3}c|*{4}c|*{4}c|}
 \multicolumn{4}{l}{\phantom{$I^I$}} &\multicolumn{4}c{Noisy Images} &\multicolumn{4}c{Clean Images}\\
\hline
\phantom{$I^I$}	&RAW &Sensor	&\# of 	&$\sigma$ & PSNR & PSNR	& PSNR	&$\sigma$   & PSNR & PSNR	& PSNR\\
Camera &Image Size &Size(mm) &Scenes&Avg. &Min.&Avg.&Max.&Avg. &Min.&Avg. &Max.\\
\hline
S90\phantom{$I^I$}		&3684$\times$2760	&7.4$\times$5.6 &40	&18.25	&17.43	&26.19  &33.39	&3.07	&35.03 &38.70 &43.43\\
T3i 		&5202$\times$3465	&22.3$\times$14.9 	&40	&11.71	&18.94	&27.44  &35.26	&2.57	&34.98	&40.43	&48.13\\
Mi3		&4208$\times$3120	&4.69$\times$3.52		&40	&19.23	&12.75	&23.49	&36.68	&3.71	&33.50	&37.09	&45.25 \\	

\hline
\end{tabular}
}
\vspace{-4mm}
\end{table*}

The final difficulty we experienced came in the form of tools to help maintain a static scene. With the other cameras we used a tripod and were able to program the automatic acquisition of the scene. With the mobile phone camera we had to use a small phone tripod and a bluetooth mouse to preserve the static scene when taking the images manually.  

Settling on the Xiaomi Mi3 phone we collected 6 images per scene. The first two images were both low-noise images and these images were averaged and set as the reference image in the alignment process. Similarly, the last two images were also both low-noise images and the last two images as well were averaged and used in the overall PSNR computation of the 'sandwich' procedure. If any movement or saturation was detected the images were cropped appropriately post alignment. In the end many of the scenes for the Mi3 were cropped, but all are static with PSNR around 35 or more.

\subsection{Main Assumptions and Notations}\label{sec:assumptions}

In this section we present the main assumptions that form the basis of the acquisition procedure, intensity alignment, and noise level estimation.

The following notations will be used in this paper:
\begin{itemize}
\item $R,I^r$ -- the reference image
\item $C,I^c$ -- the clean image
\item  $N,I^n$ -- the noisy image
\item $GT,I^{GT}$ -- the unknown ground truth image
\item $\epsilon,\epsilon_r,\epsilon_c$ -- random variables for noise in the low-noise images
\item $\epsilon_n$ -- random variable for noise the noisy images
\item $\sigma^2(X)=\var(X)$ the variance of a random variable $X$
\end{itemize}
 We assume that the two low-noise images $I^r$ (reference)  and $I^c$ (clean) as well as the noisy image(s) $I^n$ (acquired with the "sandwich" procedure from Section \ref{sec:acq}) are all noisy versions of a common (unknown) ground truth image $I^{GT}$, corrupted by zero-mean noise. Thus:
\begin{equation}
\begin{split}
I^n(x)&=I^{GT}(x)+\epsilon_n(x)\\
I^r(x)&=I^{GT}(x)+\epsilon_r(x)\\
I^c(x)&=I^{GT}(x)+\epsilon_c(x) \label{eq:noise}
\end{split}
\end{equation}
where $\epsilon_n(x),\epsilon_r(x)$, and $\epsilon_c(x)$ are zero-mean and independent of each other.

 We also assume that the reference and clean images have the same noise distribution since the two images are of the same static scene with the same ISO level and exposure time. Note that the reference and clean images have low amounts of noise because many photons have been accumulated on the sensor during the long exposure time. 

 In summary our assumptions are:
 \begin{enumerate}
 \item The images are formed as described in eq. \eqref{eq:noise} with 
 \[E[\epsilon_n(x)]=E[\epsilon_r(x)]=E[\epsilon_c(x)]=0.\]
 \item  For any $x$, the random variables $\epsilon_n(x),\epsilon_r(x), \epsilon_c(x)$ are independent.
 \item For any $x$, $\epsilon_r(x)$ and $\epsilon_c(x)$ are identically distributed.
 \end{enumerate} 
 
 It is shown in \cite{NoiseEstimation} that the noise in the digital camera images has short range correlations. 
 We don't need to make any assumptions about the spatial correlations inside one image, just between the three images at the same location.
 
 We will see in experiments that our estimation method based on these assumptions works very well in estimating the noise level in images.
\begin{figure*}[t]
\centering
\hskip -1mm
\includegraphics[width=4.cm, height=4cm]{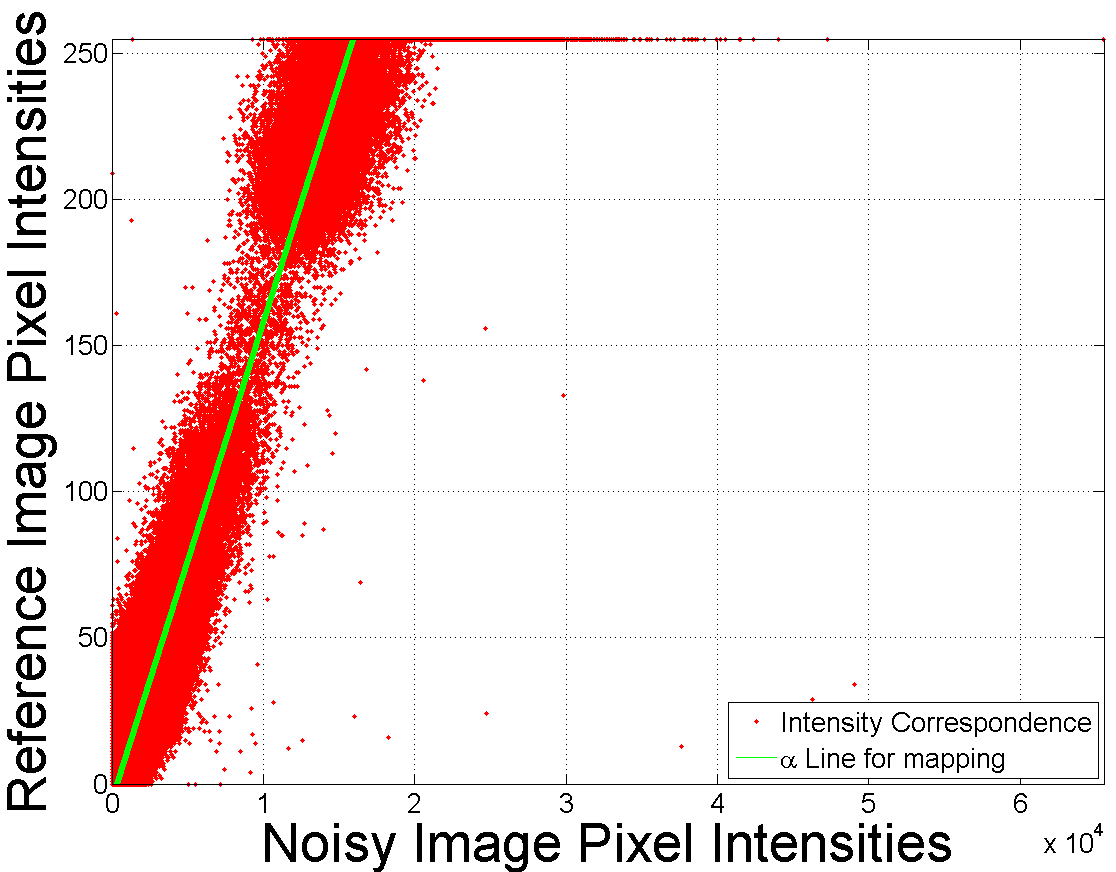}
\hskip -1mm
\includegraphics[width=4.cm, height=4cm]{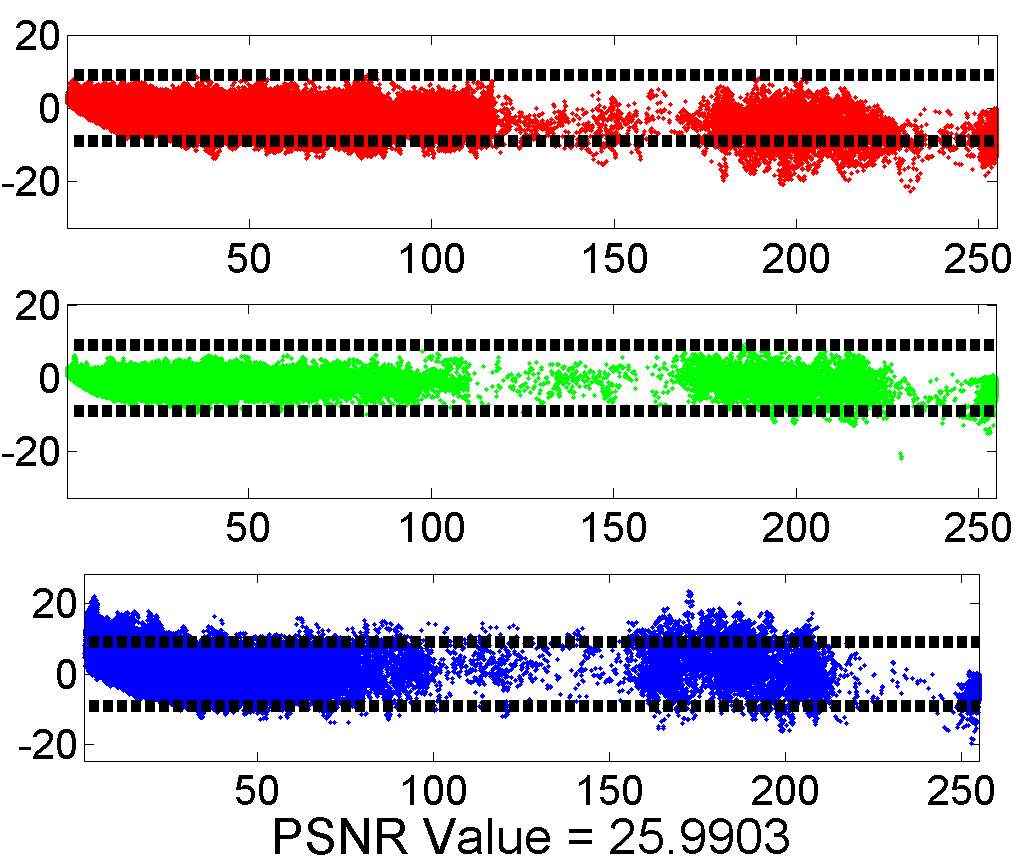}
\hskip -1mm
\includegraphics[width=4.cm, height=4cm]{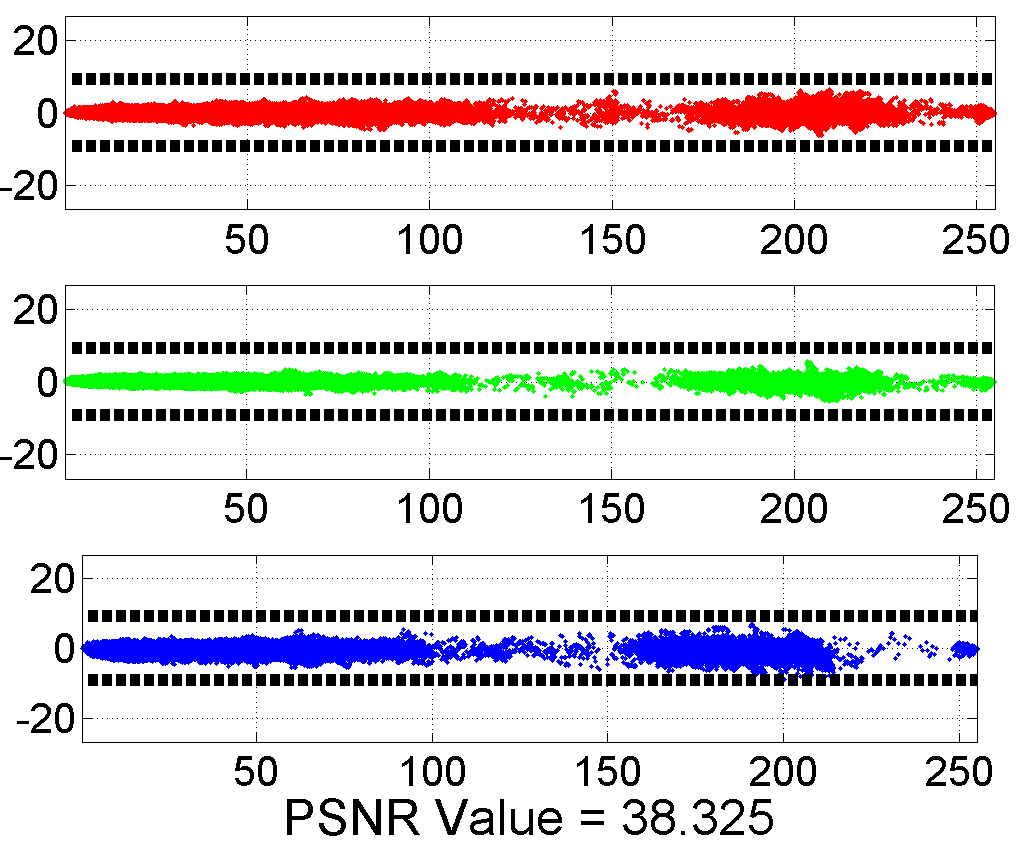}
\vskip -3mm
\caption{Scatter plots of the pixel intensity correspondence between a reference image and its noisy counterpart. Left: the correspondence between the red channel of the 8-bit reference image and the red channel for the 16-bit noisy image. The line shows the estimated linear mapping to align the noisy image to the reference image. Middle: the difference between corresponding pixel intensities of the reference 8-bit and aligned noisy 8-bit image vs reference image intensities for all three color channels. Right: the difference between corresponding pixel intensities between the reference 8-bit and the aligned clean 8-bit image vs reference image intensities. }
\label{fig:pixelcorrespondence}
\vspace{-4mm}
\end{figure*}
\subsection{Intensity Alignment} \label{sec:alignment}

The dataset construction went beyond just the acquisition of the images. For the purpose of properly aligning the pixel intensities of the image pairs, we developed a new form of brightness adjustment that mapped our RAW images to an 8-bit uncompressed format. 

The reference image was first mapped from 16-bit to 8-bit as follows. We computed the cumulative distribution of the 16-bit pixel intensities of the RAW reference image and constructed a linear scaling of the RAW reference image that sets the 99th percentile value to the intensity value 230 in the 8-bit image. Thus 1\% of the pixels are mapped to intensities above 230, and even fewer will be saturated to value 255. We chose the value 230 so that most of the noisy images will not have much saturation after alignment with the reference image.

Each of the other images of the same scene is at the same time reduced to 8-bit and aligned to the 8-bit reference image by finding a linear mapping specified by parameter $\alpha$ such that if $I$ is the 16-bit image, the 8-bit aligned image is obtained from $\alpha I$ after its values larger than 255 or less than 0 are truncated. For better accuracy,  instead of working with the two images $I$ and $R$, we use blurred versions $\tilde I$ and $\tilde R$ obtained by convolution with a Gaussian kernel with $\sigma=5$ to estimate the intensity alignment parameter $\alpha$. This way the level of noise is reduced. To avoid biases obtained near intensity discontinuities, the alignment parameter is computed based on the low gradient pixels $M=\{i, |\nabla \tilde R(i)|<1\}$.

The parameter $\alpha$ is found to minimize
\[
E(\alpha)=\sum_{i\in M} (\tilde R(i)-\max[\min(\alpha \tilde I(i),255),0])^2
\]
\begin{figure}[t]
\centering
\includegraphics[height=4cm,width=4.5cm]{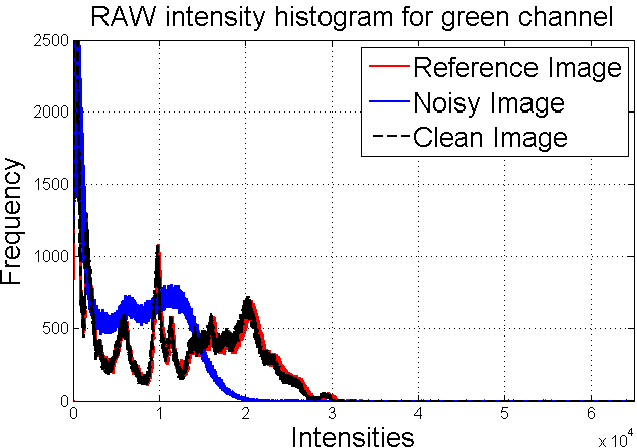}
\includegraphics[height=4cm,width=4.5cm]{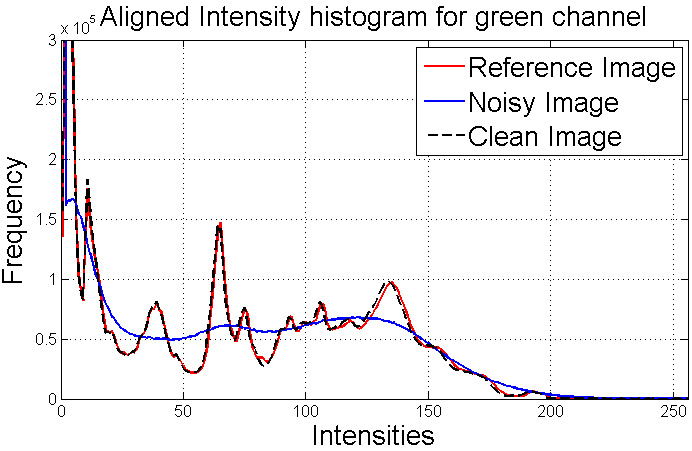}
\vskip -4mm
\caption{An example of the pixel intensity histogram for the Clean and Noisy Green Channels before and after our brightness alignment. }
\label {fig:alignment}
\vspace{-3mm}
\end{figure}

This is done by coordinate optimization using the Golden section search in one dimension \cite{press2007numerical} optimizing on $\alpha$ until convergence. The parameter $\alpha$ obtained for the mapping is robust to outliers. 

In Figure \ref{fig:pixelcorrespondence}, left is shown an example of the correspondence between the pixels of the 8-bit reference image $R$ and the RAW noisy image $I$ of the same scene, with the alignment line parameterized by $\alpha$ superimposed. The alignment of the obtained 8-bit image $I$ with the 8-bit reference can be diagnosed by plotting the intensity difference $\tilde I-\tilde R$ (blurred versions) vs the reference intensity $\tilde R$. This is shown in Figure \ref{fig:pixelcorrespondence}, middle for the noisy image and Figure \ref{fig:pixelcorrespondence}, right for the clean image.  We obtained plots like Figure \ref{fig:pixelcorrespondence} for all the images in the dataset as a way of diagnosing any misalignment or nonlinear correspondence between the reference image and the corresponding noisy or clean images. The dark dashed horizontal line in the middle and right plots are 95\% noise bounds for clean images with at least PSNR = 35. Figure \ref{fig:alignment} shows an example of the green channel for a particular image in the dataset before and after alignment is performed. 

\subsection{Noise Estimation}\label{sec:noiseeq}

As stated previously the amount of noise present in the dataset is due to the sensor and the amplification process. The fact that not all of the images were taken in the same environment under the same camera settings means that we have a wide variety of noise in our images. The fact that we are not dealing with artificial noise  also means that we do not know beforehand what will be the noise variance $\sigma^2$. Thankfully our "sandwich" procedure for image acquisition, as influenced by \cite{RadiometricCCD,NoiseEstimation}, allows us to estimate the noise level for any of our images. The noise level can be estimated locally for an image patch or globally for the entire image.

We will use the fact that if two random variables $A,B$ are independent, then $\var(A-B)=\var(A)+\var(B)$, or in other words $\sigma^2(A-B)=\sigma^2(A)+\sigma^2(B)$ where $\var(A), \sigma(A)$ are the variance and standard deviation of A respectively.
Then  from equation \eqref{eq:noise} we get
\[\sigma^2(I^r(x)-I^c(x))=\var(\epsilon_r(x)-\epsilon_c(x))=\var(\epsilon_r(x))+\var(\epsilon_c(x))=2\sigma^2(\epsilon(x))\]
from the independence of $\epsilon_r(x)$ and $\epsilon_c(x)$ and the fact that $\epsilon_r(x)$ and $\epsilon_c(x)$ are identically distributed (so we can represent them as $\epsilon(x)$). We obtain the estimation of the noise level in the clean and reference images:
\begin{equation}
\sigma^2(I^r(x)-I^{GT}(x))=\sigma^2(I^c(x)-I^{GT}(x))=\sigma^2(\epsilon(x))=\frac{1}{2}\sigma^2(I^r(x)-I^c(x)) \label{eq:noiseclean}
\end{equation}

For the noisy images we use 
\[\sigma^2(I^n(x)-I^r(x))=\var(\epsilon_n(x)-\epsilon_r(x))=\sigma^2(\epsilon_n(x))+\sigma^2(\epsilon_r(x))\]
to obtain the estimation of the noise level as
\begin{equation}
\sigma^2(\epsilon_n(x))=\sigma^2(I^n(x)-I^{GT}(x))=\sigma^2(I^n(x)-I^r(x))-\frac{1}{2}\sigma^2(I^r(x)-I^c(x)) \label{eq:noiselevel}
\end{equation}
\begin{figure*}[t]
\centering
\hspace{-3mm}\includegraphics[height=4.cm,width=4.1cm]{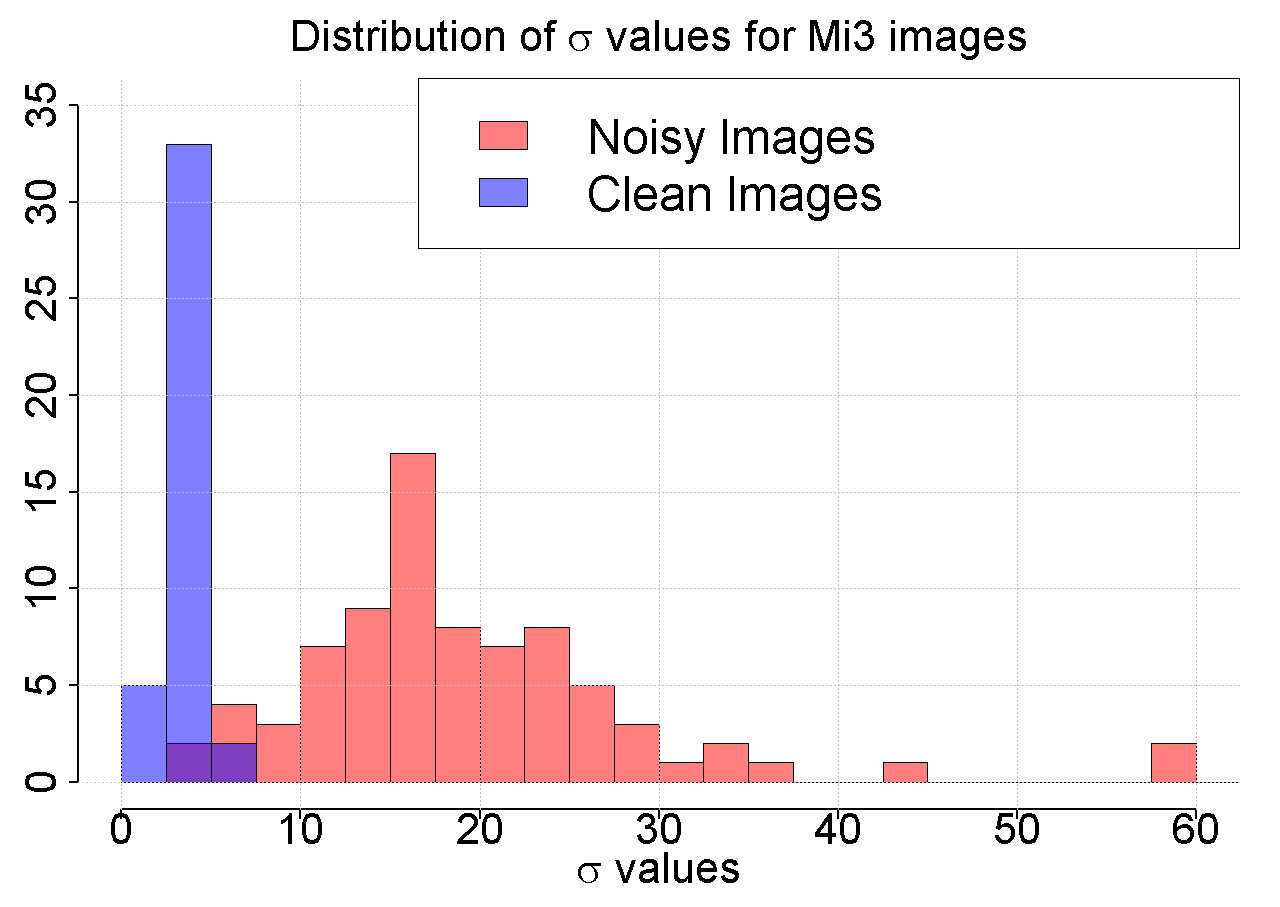}
\hspace{-3mm}\includegraphics[height=4.cm,width=4.1cm]{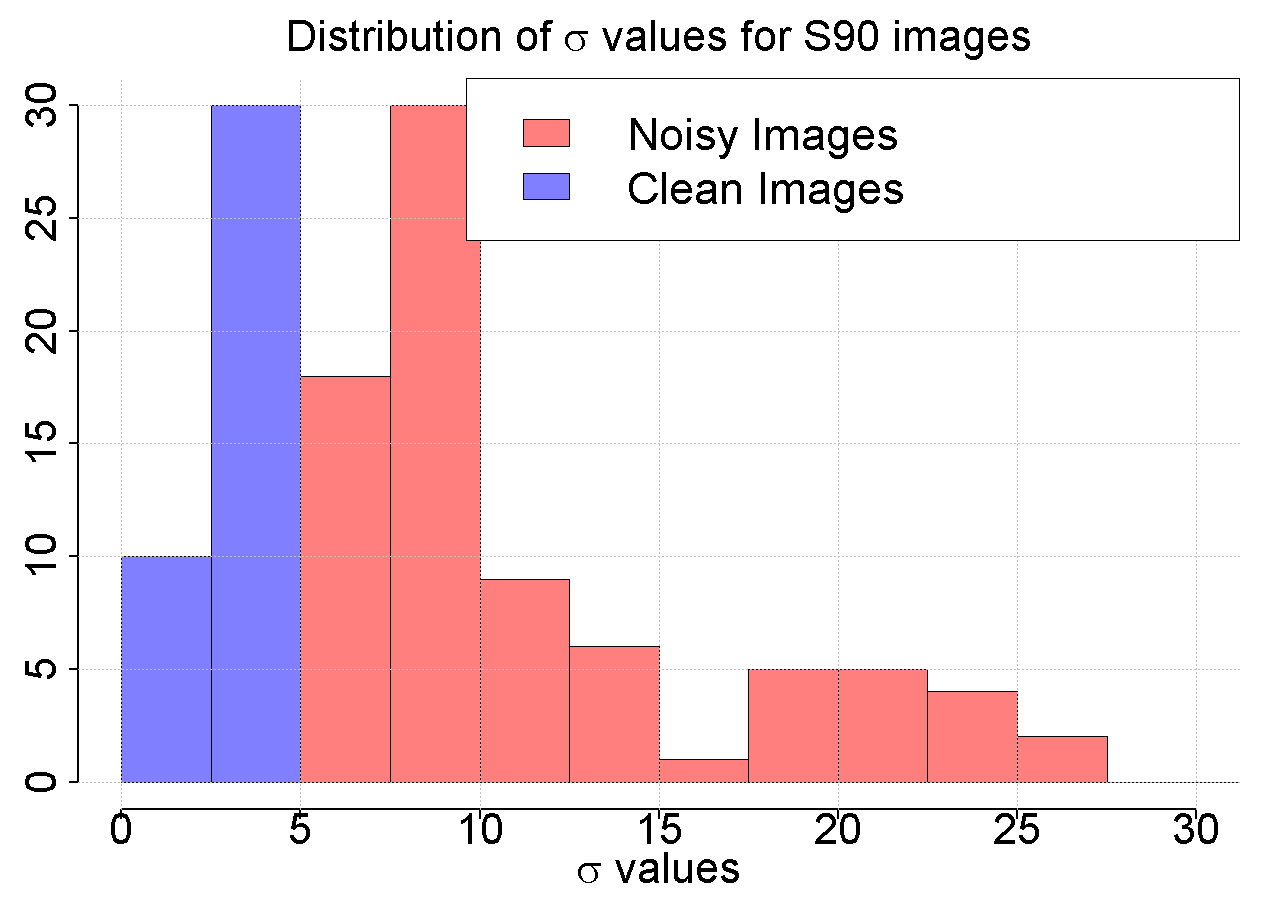}
\hspace{-3mm}\includegraphics[height=4.cm,width=4.1cm]{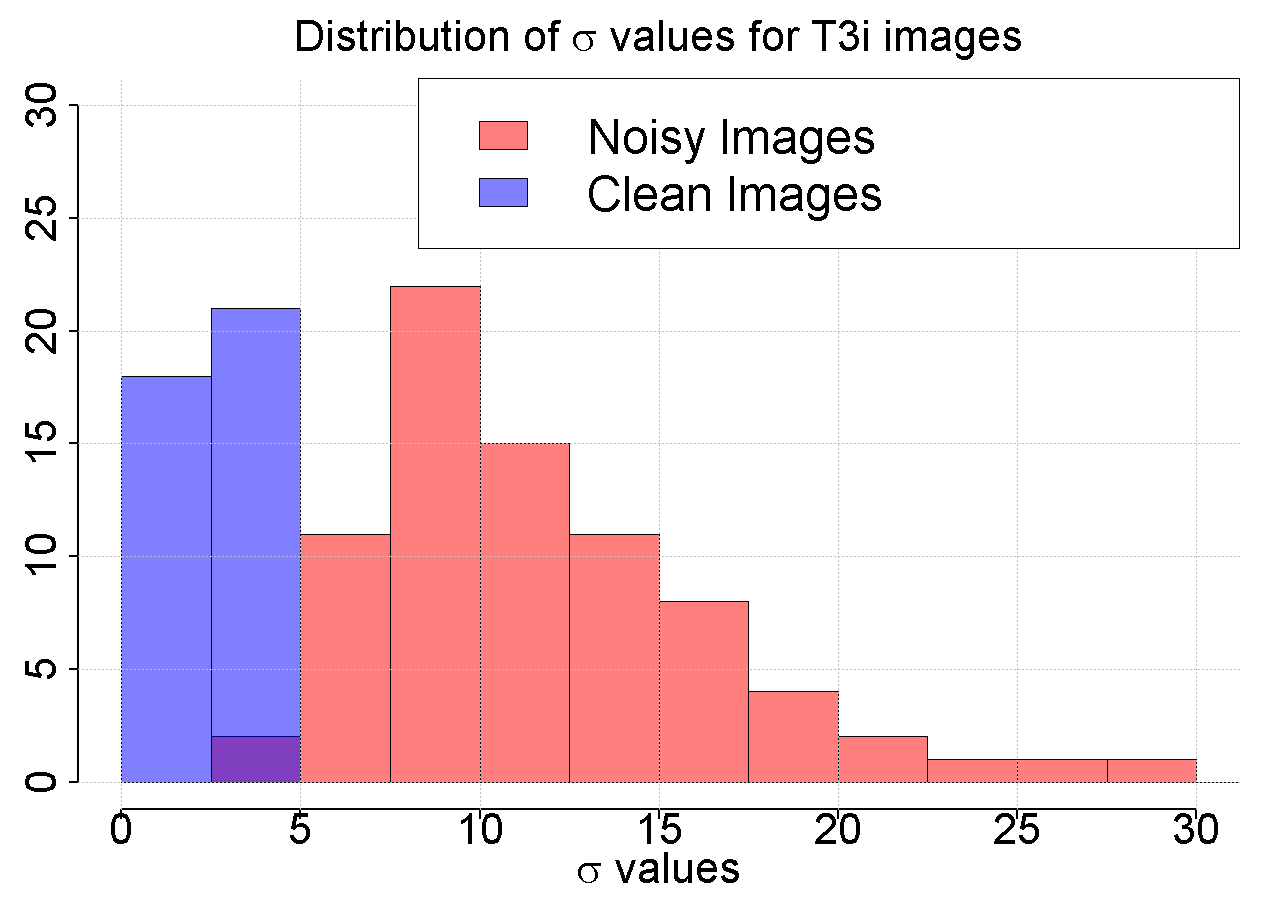}
\vskip -4mm
\caption{Frequency distributions of various noise levels for the noisy and clean images obtained from the Mi3, S90, and T3i  cameras respectively. }
\label{fig:hists}
\vspace {-4mm}
\end{figure*}

If we want to use the best estimate of the GT, which is $I^a(x)=(I^r(x)+I^c(x))/2$, then we have an alternative formula for the noise level in the noisy images
\begin{equation}
\sigma^2(\epsilon_n(x))=\var(I^n(x)-I^{GT}(x))=\var(I^n(x)-I^a(x))-\frac{1}{4}\var(I^r(x)-I^c(x)) \label{eq:noiselevel1}
\end{equation}

We can use equations \eqref{eq:noiseclean} and \eqref{eq:noiselevel} to estimate the true noise level for any image in our dataset. Again, these noise levels can be computed globally for the whole image or locally on a patch basis.



\section{Dataset Information}

Aside from estimating the noise level in every image, we also quantified the image fidelity across the image batches using various metrics such as PSNR \cite{perceptualdistortion}, SSIM \cite{ssim}, and VSNR \cite{vsnr}. In particular we modified the PSNR measurement by incorporating our estimate of the noise from \eqref{eq:noiselevel} as opposed to using the standard noise estimate from the difference image between a clean and noisy image pair.  Although there exist specialized metrics for low-light conditions such as \cite{similaritylowlight} we decided to use measures that are the most prevalent and common in practice.

\begin{figure}[t]
\vspace{-3mm}
\centering
\includegraphics[width=4.1cm,height=4cm]{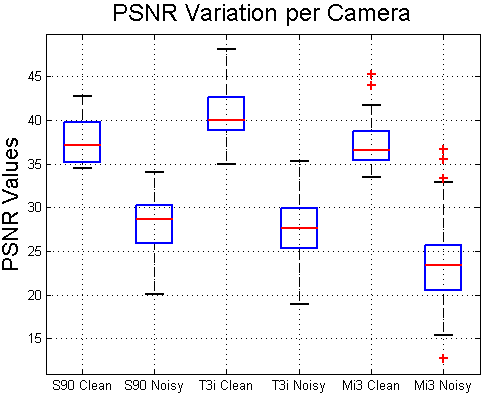}
\includegraphics[width=4.1cm,height=4cm]{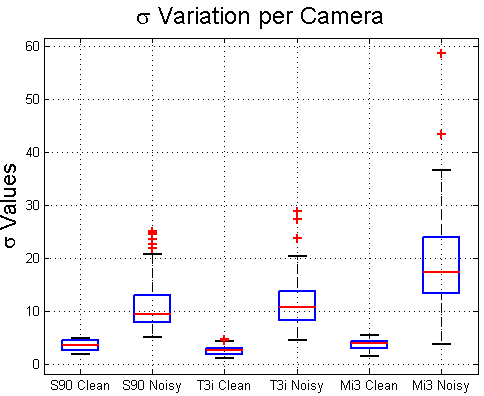}
\vskip -3mm
\caption{Variation of PSNR and $\sigma$ values for noisy and clean images for each camera.}
\label{fig:boxplots}
\vspace{-5mm}
\end{figure}

Table \ref{tab:datastats} lists some specific characteristics about the various cameras and their images in the dataset. Note that the $\sigma$ in Table \ref{tab:datastats} comes from the estimates from equations \eqref{eq:noiseclean} and \eqref{eq:noiselevel}. Figure \ref{fig:hists} shows the distribution of noise levels for the noisy and clean images for each camera. Figure \ref{fig:boxplots} shows box-plots of the variation in PSNR and noise levels for each camera.

One interesting observation that we draw from our dataset is that the low noise images still have a noise level $\sigma$ of about $3$ and as high as $5$, which is invisible to the eye. It tells us something about the local nature of the manifold of natural image patches, that the manifold is ``thick'' in the sense that perturbing a patch with (probably even Gaussian) noise of $\sigma\leq 5$ obtains another natural image patch. Such information could be useful in the study of natural image statistics or for learning generative models from natural images (e.g. autoencoders).

\section{Experiments}

In this section we will evaluate the accuracy of our noise estimation procedure and compare it to the standard noise estimate (standard deviation of the difference image) for both synthetic and real noise data. We will also compare our noise estimation framework to the Poisson-Gaussian noise model and estimation procedure presented in \cite{Foi-Poisson}. Afterwards we will examine the denoising performances of four algorithms on our dataset using three image fidelity metrics.
 
\subsection{Evaluation of Alignment and Noise Estimation Using Artificial Noise} \label{subsec:noiseeval}

\begin{figure}[htb]
\centering
\vspace{-3mm}
\hspace{-1mm}\includegraphics[width=11cm]{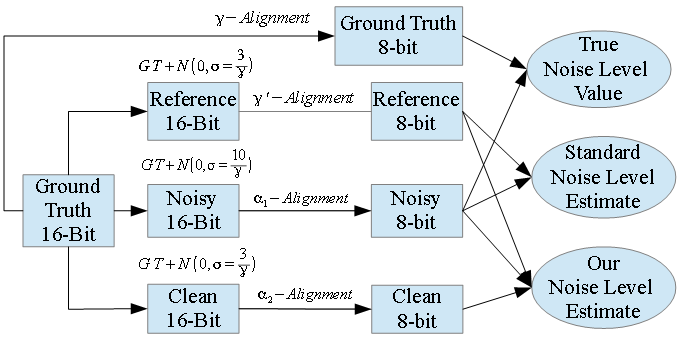}
\vskip -4mm
\caption{The process for constructing the proper reference, clean, noisy, and ground truth images necessary for the noise estimation evaluation. The values of $\gamma$', $\alpha_{1}$, and $\alpha_{2}$ represent the usual alignment of those respective images from 16-bit to 8-bit as described in Section \ref{sec:alignment}.}
\vspace{-4mm}
\label{fig:errorfc}
\end{figure}

To evaluate our intensity alignment and noise estimation method we constructed scenes with added artificial noise, as illustrated in Figure \ref{fig:errorfc}. For this we chose ten 16-bit RAW reference images from each of the three digital cameras and used them as ground truth images $I^{GT}$ for constructing artificial sequences from them. We then used our alignment method as described in Section \ref{sec:alignment} to construct an 8-bit version of $I^{GT}$. We then generated $I^r$,$ I^n$, and $I^c$ by adding artificial Gaussian noise to the 16-bit $I^{GT}$. For 16-bit $I^r$ and $I^c$ we added $\sigma=\frac{3}{\gamma}$ amount of noise where $\gamma$ is the multiplication factor to map the 16-bit $I^{GT}$ to an 8-bit $I^{GT}$. A 16-bit $I^n$ was generated using $\sigma=\frac{10}{\gamma}$. This way the standard deviation of the difference from the 8-bit $I^{GT}$ to the 8-bit  $I^r$ (or  $I^c$) will be 3 and to the 8-bit  $I^n$ it will be 10.
 We then performed our standard alignment on $I^r$,$ I^n$, and $I^c$ to map them over to 8-bit images obtaining parameters $\gamma',\alpha_i,\alpha_2$, as illustrated in Figure \ref{fig:errorfc}. 

Observe that multiplying the alignment parameters $\gamma',\alpha_i,\alpha_2$ by the same factor produces another alignment of the same quality, thus the alignment is only identifiable up to a multiplicative constant. For this reason, the alignment is evaluated indirectly, through the quality of the noise level estimation. 

The true values of the noise levels for the $I^r$,$ I^n$, and $I^c$ images can be computed as the standard deviation of the difference between each of them and $I^{GT}$ (all in 8-bit versions). 
Our noise level estimation method for $I^c$ and $ I^n$ described in Section \ref{sec:noiseeq} is compared with the true noise level to obtain the relative error  (defined as estimation error divided by the true noise level value). The same type of relative error is also computed for the standard estimate of the noise, which is $\sigma(I^n-I^r)$ and $\sigma(I^c-I^r)$.

\begin{figure}[t]
\vspace{-2mm}
\centering
\includegraphics[width=5.1cm,height=5cm]{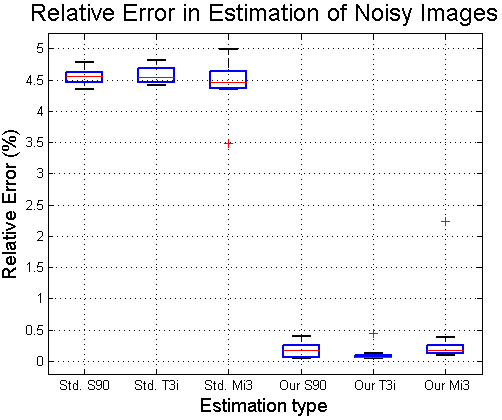}
\includegraphics[width=5.1cm,height=5cm]{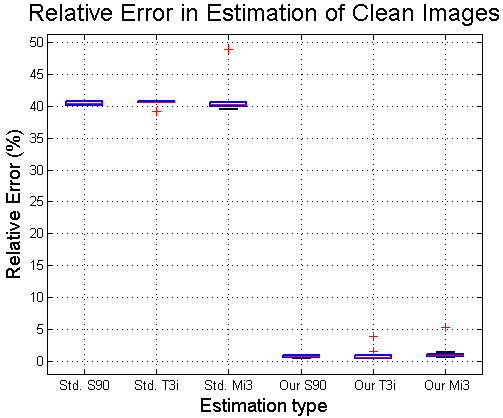}
\vskip -3mm
\caption{The relative error in estimating noisy and clean images. }
\label{fig:errorbp}
\vspace{-4mm}
\end{figure}

Figure \ref{fig:errorbp} shows the relative error (defined as error divided by the true value) of estimating the noise level for both $I^n$ and $I^c$ . When it came to estimating the noise level $I^n$, our estimation method kept the relative error to below $0.5\%$, while the standard method of estimating the noise level had a relative error around $5\%$. For the low noise images $I^r$ and $I^c$ our method had an error below 1\% while the standard method had an error of around 40\%.

For all but four of the 90 images evaluated, the relative estimation error for our alignment and noise estimation method was below 1\%. This gives us confidence that our intensity alignment method together with the proposed noise level estimation method provide an accurate estimation of the true noise level in images, at least on data with artificial noise.


\subsection{Evaluation of Noise Estimation Using Real Noise}

\begin{figure}[htb]
\vskip -4mm
\centering
\includegraphics[width=6.cm]{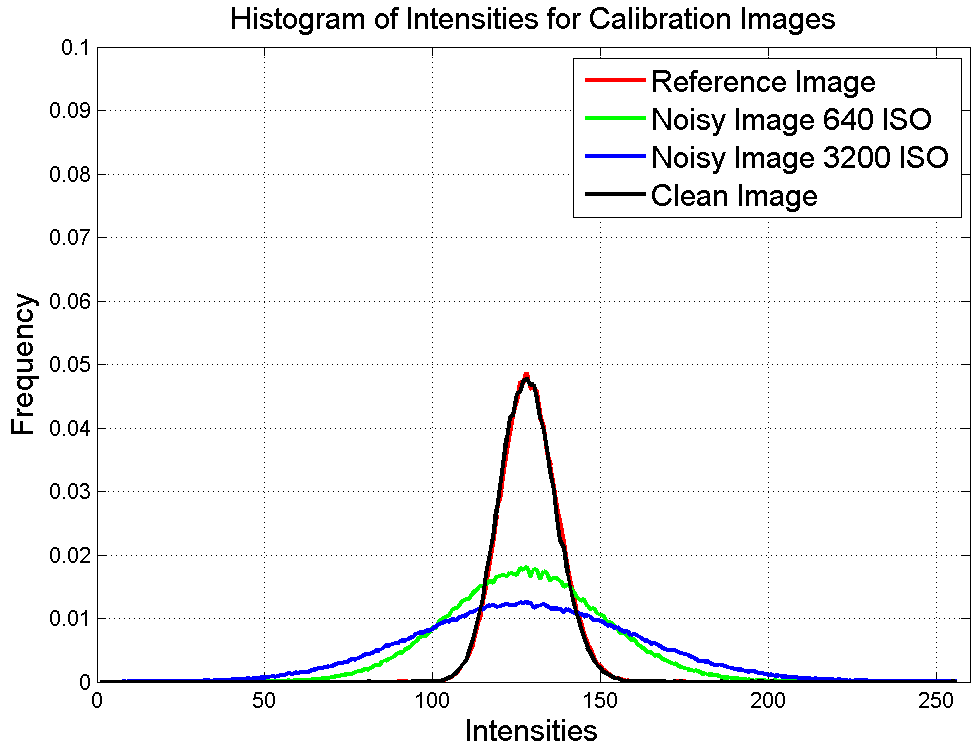}
\includegraphics[width=6.cm]{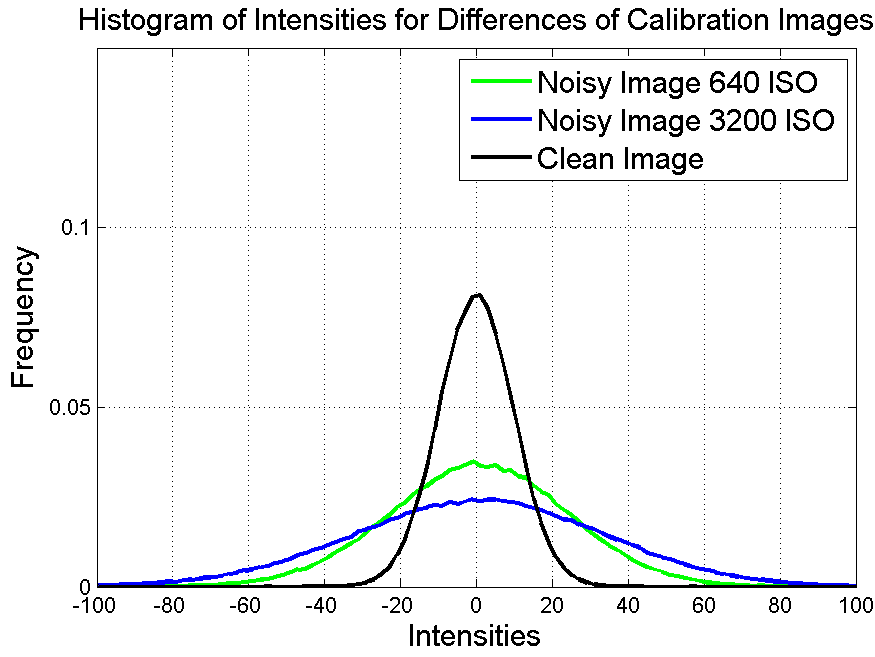}
\vskip -3mm
\caption{Analysis of the calibration images. Left: the intensity histograms of the green channels of the calibration images. Right: the distribution of the intensity difference between the reference image and the various other images in the calibration dataset. \label{fig:Calibration}}
\vspace{-4mm}
\end{figure}

To further investigate how well the assumptions we made in Section \ref{sec:assumptions} about the noise hold, we acquired a special scene with the S90 camera. The scene was of a constant intensity surface in low-light settings. Using our intensity alignment methodology, instead of mapping our clean image from the 99th quantile to intensity 230, we mapped the median to intensity 128. Using this mapping we then aligned the other two noisy and the clean image using the Golden section method, as described in Section \ref{sec:alignment}. Figure \ref{fig:Calibration} shows the alignments of the calibration dataset as well as a histogram of pixel difference between the reference image and the other images in the calibration dataset.  

\begin{figure}[htb]
\vspace{-4mm}
\centering
\includegraphics[width=6.5cm,height=5cm] {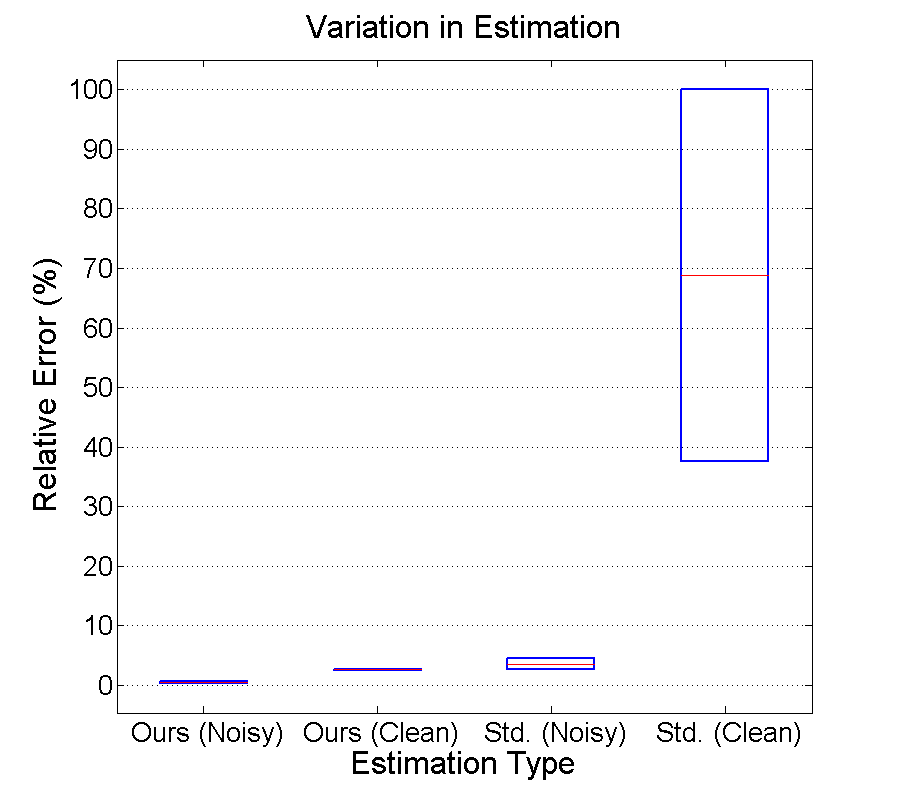}
\vskip -5mm
\caption{Comparison between our method of estimating $\sigma$ and the standard method based on the difference image. Both methods were tasked with estimating the $\sigma$ for the red, green, blue channels, and the overall image for the calibration scene.
 \label{fig:CalibrationSigma}}
\vspace{-4mm}
\end{figure}

Because we know that the $I^{GT}$ was constant since the scene contained a constant intensity surface, we can immediately obtain a true value for ${\sigma}^2$ for each image by directly computing the intensity variance in the image. However, to account for smoothly changing illumination, we constructed a GT version for each image by Gaussian blurring it with a large spatial kernel ($\sigma=20$) and then calculated the noise level as the variance of the difference image between the original image and its smoothed version. We then looked to see if the standard estimate of using the difference image between the reference image and the other calibration images provided similar results to those we obtained using our methodology from equations \eqref{eq:noiseclean} and \eqref{eq:noiselevel}. Analysis of the estimated noise levels for the three image channels and the overall estimate are summarized with boxplots in Figure \ref{fig:CalibrationSigma}. 

As Figure \ref{fig:CalibrationSigma} shows, our estimated $\sigma$ values are less biased and have smaller variance than the standard estimation of $\sigma$ from the difference images. The average relative error for our method of estimation is $1.58\%$ and for the standard method of estimation is $36.22\%$. The results that we obtained for this evaluation are in line with the results we obtained for noise estimation for images with artificial noise. Thus our investigation gives us enough confidence in our estimation going forward. Consequently, the noise estimation described in Section \ref{sec:noiseeq} will be used as our noise estimation method for all of the images in our dataset and for estimating the PSNR of the denoised images.

\subsection{Evaluation of the Poisson-Gaussian Noise model} \label{sec:poisson}

As stated previously, the Poisson-Gaussian noise model \cite{Foi-Poisson,makitalo2014noise} depends on the image intensity. With our notation introduced in Section \ref{sec:assumptions}, the observed image intensity $I^n(x)$ is represented under the Poisson-Gaussian noise model as
\begin{equation}
I^n(x)=\alpha p(x)+n(x)
\end{equation}
where $p(x)$ is an independent Poisson random variable with expected value $y(x)=I^{GT}(x)/\alpha$ and $n(x)$ is an i.i.d. Gaussian $n(x)\sim N(0,\tau^2)$. This way we obtain the noise model
\begin{equation}
\epsilon_n(x)=I^n(x)-I^{GT}(x)=\alpha p(x)+n(x)-I^{GT}(x)
\end{equation}
which is independent and has zero mean. Therefore the Poisson-Gaussian noise model obeys the assumptions made in Section \ref{sec:assumptions}. 

Under the Poisson-Gaussian noise model the noise level (standard deviation) has an exact relationship with the noise-free image through  $\sigma(\epsilon_n(x))=\sqrt{\alpha I^{GT}(x)+\tau^2}$.

In \cite{Foi-Poisson} is presented a maximum likelihood approach for estimating of the Poisson-Gaussian model parameters $(\alpha,b)$, (where $b=\tau^2$) from a single image. The authors also observe that $b$ could also be negative due to the pedestal level, a constant offset from zero of the digital imaging sensor.

Using local noise estimation through equations \eqref{eq:noiseclean}, \eqref{eq:noiselevel}, and \eqref{eq:noiselevel1}, we can calculate the the intensity level sets $I^{GT}_i$ of each image patch and the variance $v_i$ of the noise in each level set, then we can find the Poisson-Gaussian model parameters $(\alpha,b)$ so that $v_i=\alpha I^{GT}_i+b$ by the maximum likelihood method from \cite{Foi-Poisson}. We can then see how well the Poisson-Gaussian noise model fits our data and we can we can compare our model parameters with the single-image parameter estimates from \cite{Foi-Poisson}.

A special scene was acquired for this purpose using a uniform background with a smoothly changing intensity and our "sandwich" procedure. We converted the images to gray-scale to be able to compute the model from \cite{Foi-Poisson} and divided the image into $400\times 400$ blocks and the blocks into intensity level sets of a smoothed image, following the method described in \cite{Foi-Poisson}. Based on the different ways to estimate the noise variance $\sigma$ in each level set 
we considered the following variants:
\begin{itemize}
\item In the Foi model, the noise level $\sigma$ is estimated as the standard deviation of the wavelet detail coefficients $z^{wdet}$, as described in \cite{Foi-Poisson}.
\item In the three image model, the noise level $\sigma$ is estimated using three images (reference, clean and noisy) and equations \eqref{eq:noiseclean} and \eqref{eq:noiselevel1}. 
\item In the blurred reference model, the $I^{GT}$ is obtained by blurring the reference image with a large Gaussian kernel and the noise level $\sigma$ is estimated as the standard deviation of the difference between the noisy image and the blurred reference image. 
\item The blurred noisy model takes as $I^{GT}$ the blurred noisy image, thus it is obtained entirely from the noisy image.
\end{itemize}
The smooth image for obtaining the level sets was obtained as follows: for the Foi method we used the blurred $z^{wapp}$ wavelet approximation image, for the three image model we used the average $(I^r+I^c)/2$ between the reference and clean images, for the blurred reference model we used the blurred reference image, and for the blurred noisy model we used the blurred noisy image.

On the obtained $(\hat y_i,\hat \sigma_i)$ intensity-noise level pairs for each variant we fitted Poisson-Gauss model parameters $(\alpha,b)$ by maximum likelihood as described in  \cite{Foi-Poisson}.

To validate our implementation of the Poisson-Gaussian model estimation procedure, we also plotted the Foi original model, where the parameters were obtained from the noisy image using code available online\footnote{Code obtained from \url{http://www.cs.tut.fi/~foi/sensornoise.html}}.
\begin{figure*}[t]
\centering
\includegraphics[width=6.cm,height=6cm]{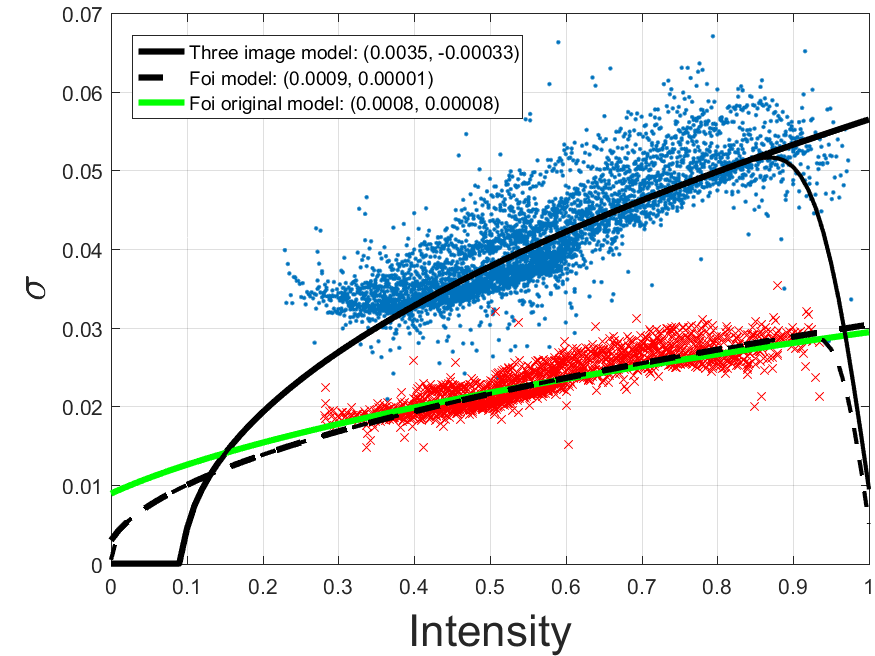}
\includegraphics[width=6.cm,height=6cm]{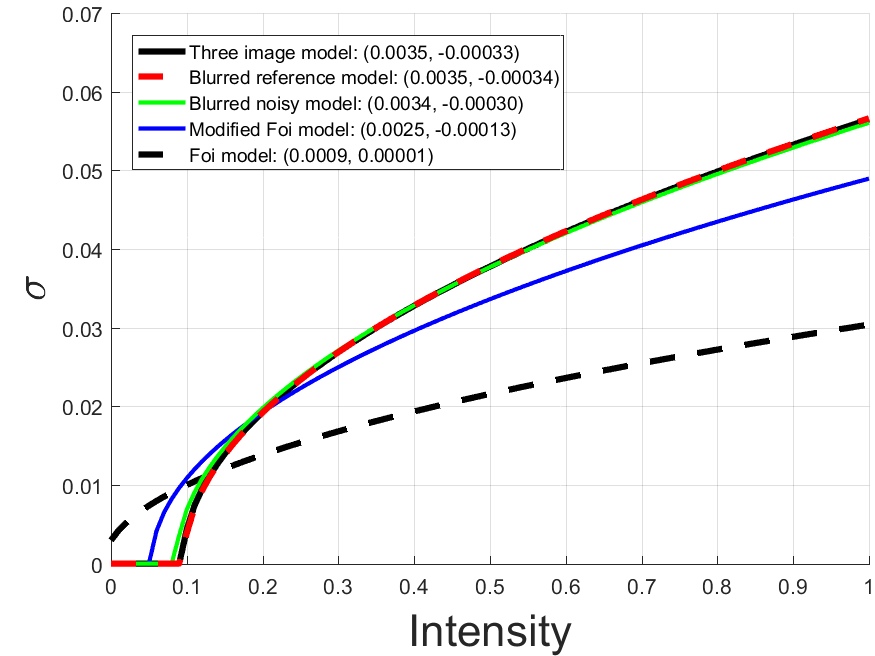}
\vskip -5mm
\caption{Left: The noise curve of our pair image model and the Foi Poisson-Gaussian Mixture model. Right: The noise curves for various noise estimation models using image pairs and the Foi Poisson-Gaussian Mixture model.}\label{fig:sigmaplots}
\vspace{-3mm}
\end{figure*}

The obtained curves are shown in Figure \ref{fig:sigmaplots}. In Figure \ref{fig:sigmaplots}, left are also shown the data points $(\hat y_i,\hat \sigma_i)$ from which the three image model and the Foi model were obtained.

 In Figure \ref{fig:sigmaplots}, right are shown the the three estimation methods based on our data: the three image mode, the blurred reference model and the blurred noisy model. Observe that these curves are very close to each other and that the blurred reference model exactly coincides with the three image model. This shows that the blurred reference is a very good approximation of $I^{GT}$ in this case and the blurred noisy image is also a good approximation of $I^{GT}$. Observe that out of these three methods only the three image model directly generalizes to images with edges, while the other methods need to carefully avoid pixels close to the edges where the blurred image does not approximate the $I^{GT}$ correctly.
 
 It can be immediately noted that the Foi estimation method is consistently below our three curves. At high intensities (around 0.9 on the normalized scale), Foi's estimate and the corresponding data points are underestimating the noise level by about 40\%.  
We suspect that this is due to the local correlations in the image noise, which interfere with the noise estimation based on wavelets and a simple convolution. To investigate this further, we modified the Foi model by estimating the noise level $\sigma$ as the standard deviation of the image difference $z^{diff}=I^{n}\downarrow 2-z^{smo}$,  where $I^{n}\downarrow 2$ means the noisy image downsampled by a factor of $2$. This curve can also be seen in Figure \ref{fig:sigmaplots}, right. Compared to Foi's original model, this modified Foi model brings the curve quite close to our estimates. It is still not exactly the same because the $z^{smo}$ is not a good approximation of the $I^{GT}$ since it was obtained by smoothing with a box kernel which does not remove the noise as well as a Gaussian kernel. Indeed the blurred noisy model, which does the smoothing of the noisy image with a large Gaussian kernel to obtain the $I^{GT}$ and level sets, comes very close to our estimates.
  
Foi's Poisson-Gaussian noise model is using just the noisy image to infer the noise level in the image. With our data acquisition methodology we have both clean and noisy images and are able to infer more accurately the noise level in the image. Also, note that this evaluation was done on a special scene of a uniform background of continuously changing intensity and no edges. Foi's estimation method would have more difficulty in estimating the noise curve in images with a lot of edges or a lot of textures. At the same time, the three image approach only used the aligned images and no blurring, so it should work as well when edges are present.

\subsection{Evaluation of Denoising Algorithms}

In this section we use our dataset to evaluate six popular image denoising algorithms: the Active Random Field (ARF)\cite{barbu2009training}, Block Matching and 3D Filtering (BM3D) \cite{dabov2007image}, Bilevel optimization (opt-MRF) \cite{optMRF}, Multi Layer Perceptron (MLP) \cite{NeuralNetworkCompete}, Non-local means using a James-Stein estimator(NLM-JS) \cite{JS-NLM}, and Non-local means with a soft threshold (NLM-ST) \cite{lu2015non}. These algorithms were selected because they are efficient enough to handle our large images and have code available online. Each of these methods depends on a noise level parameter $\sigma$. We tested the ARF filters \footnote{from \url{http://ani.stat.fsu.edu/~abarbu/ARF/demo.zip}} that were trained using Gaussian noise (in particular the trained filters for $\sigma$ = 10, 15, 20, 25 and 50) and using four iterations. A special version of the BM3D algorithm meant for color image denoising \footnote{from \url{http://www.cs.tut.fi/~foi/GCF-BM3D/BM3D.zip}} was used on the noisy images. For BM3D we evaluated the algorithm's performance at $\sigma$ = 5, 10, 15, 20, 25, and 50. For opt-MRF we used the Gaussian trained filters ($\sigma$ = 15 and 25)  and a maximum limit of 30 iterations for the optimization \footnote{from \url{http://gpu4vision.icg.tugraz.at/index.php?content=downloads.php}}. We also used MLPs trained on Gaussian filters 
\footnote{from \url{http://people.tuebingen.mpg.de/burger/neural_denoising/}} to denoise our images. In particular we used filters for $\sigma$ =10, 25, 35, 50, and 75. These methods were evaluated for the values of the noise level $\sigma$ given above and then  for each method the parameter was fixed to the value that gave the best results. 
For the NLM-JS \footnote{from\url{https://www.mathworks.com/matlabcentral/fileexchange/40162-james-stein-type-center-pixel-weights-for-non-local-means?s_tid=srchtitle}} algorithm a patch size of $3 \times 3$, with a search window of $15 \times 15$, and block size of $15 \times 15$ was used for denoising. Finally, for the NLM-ST \footnote{from\url{http://ieeexplore.ieee.org/document/6957527/media}} algorithm a patch size of $5 \times 5$, with a search window of $13 \times 13$, and block size of $21 \times 21$ was used for denoising. 
The noise parameter value for the NLM-JS and NLM-ST algorithms were estimated for each image using our previously discussed "sandwich" procedure. For the ARF, opt-MRF, MLP, NLM-JS, and NLM-ST algorithms the image channels were denoised in the YUV color space for better performance. For BM3D there was no transformation necessary because it could directly handle color images.

\begin{table}[htb]
\vspace{-4mm}
\centering
\caption {Performance of various denoising algorithms on our dataset. \label{tab:DenoisingResults}}
\scalebox{1}{
\begin{tabular}{l c c c c c c c}
\hline
Camera\phantom{$I^I$}		&Before Denoising		&ARF	 &BM3D &opt-MRF	 &MLP	&NLM-JS &NLM-ST	\\
\hline
\hline
 \multicolumn{7}{l}{PSNR} \\
\hline
Mi3\phantom{$I^I$} &23.492 &30.918 &32.347 &31.641 &31.230 &31.348 &30.866 \\
S90	&26.187 &33.797 &36.752 &34.983	&34.073	&34.135 &33.076 \\
T3i	&27.442 &36.550   &39.966  &38.646  &37.584	&37.400  &36.822\\
\hline
Average\phantom{$I^I$}  &25.707 &33.755 &36.355 &35.090 &34.296 &34.294 &33.589 \\
\hline
\hline
 \multicolumn{5}{l}{SSIM} \\
\hline
Mi3\phantom{$I^I$} &0.989 &0.972 &0.982 &0.964 &0.929	&0.965 &0.970 \\
S90	&0.988 &0.959 &0.979 &0.958 &0.920	&0.973  &0.970 \\
T3i	&0.991 &0.993 &0.994 &0.993 &0.933	&0.993 &0.992 \\
\hline
Average\phantom{$I^I$}  &0.989 &0.981 &0.985 &0.972 &0.927 &0.977 &0.977\\
\hline
\hline
 \multicolumn{5}{l}{VSNR} \\
\hline
Mi3\phantom{$I^I$} &17.746 &22.387 &24.820 &22.521 &24.132 &24.201 &23.398\\
S90	&23.789 &26.769 &28.635 &27.357 &27.255 &27.282 &25.559\\
T3i	&22.318 &28.567 &30.481 &29.803 &29.429 &28.834 &28.391  \\
\hline
Average\phantom{$I^I$} &21.284 &25.908 &27.979 &26.560  &26.605 &26.828 &25.864 \\
\hline
\end{tabular}
}
\vspace {-4mm}
\end{table}


In Table \ref{tab:DenoisingResults} are shown  the denoising results of the various methods on the three cameras. 
We computed the PSNR, SSIM, and VSNR values between the denoised and the best GT estimate which is the average of the two clean images. 

Note the high values given by the SSIM prior to denoising and the lower values for two cameras after denoising. It is not clear how to interpret the SSIM results since the other two measures (PSNR and VSNR) are consistent with each other and with the fact that denoising was performed, while SSIM is not. 

The best results obtained for the ARF, opt-MRF, and MLP methods occurred with a $\sigma = 25$ filter while the BM3D provided its best results with a $\sigma =50$ filter. The results from Table \ref{tab:DenoisingResults} show that the BM3D outperformed the other methods on all the cameras using all similarity measures. In particular when comparing the performance of BM3D with MLP, opt-MRF, and NLM-ST for real noisy images these results do not lead to the same conclusions as in \cite{NeuralNetworkCompete}, \cite{optMRF}, \cite{lu2015non} where these methods performed as good or better than BM3D on small gray synthetic noisy images. 

\section{Conclusions}
In this paper we introduced a dataset of images containing real noise due to low-light settings and acquired from two digital cameras and a mobile phone. Additionally, we developed a method for obtaining pixel-aligned RAW images of low and high noise, and intensity-aligned BMP images so that proper studying of the images and their noise need not be  done in RAW format. We also presented a technique to calculate the PSNR of an image without a ground truth and we conducted extensive evaluations of our noise estimation and our alignment procedure to make sure that the difference between the noisy and clean images is just noise.

We used our data to evaluate the Poisson-Gaussian noise model  \cite{Foi-Poisson} and its parameter estimation procedure. We observed that  the noise has an overall trend that fits the Poisson-Gaussian model but the wavelet-based estimation method has difficulty estimating the correct model parameters due to short-range interactions in the noise structure and to the use of a box filter instead of a Gaussian filter for smoothing the image.

We tested our dataset on six denoising algorithms: ARF, BM3D, opt-MRF, MLP, and two version of Non-local Means. For all methods we computed the noise levels in the denoised images using a variety of methods such as PSNR, VSNR, and SSIM. Note that these denoising algorithms were trained or tuned on images corrupted by artificial Gaussian noise. Some of these methodologies (ARF opt-MRF, and MLP) and many other recent state-of-the-art denoising methods such as: CSF \cite{Shrinkagefields}, LSSC \cite{mairal2009non}, and RTF \cite{Loss-specific} learn the noise structure in a supervised way from the noisy-clean image pairs. These methods could in fact perform even better for denoising low light images if trained on our dataset. With so many different denoising methods having been developed or currently in development, our dataset allows for proper analysis of these tools, and for the quantitative evaluation of noise models for digital and mobile phone cameras. 

Our dataset poses one more training and testing challenge compared to using images corrupted by artificial noise. The images in our dataset have a large range of noise levels in them, while usually denoising methods are trained and evaluated for one known noise level only. Data with different noise levels poses many challenges in training and testing, but at the same time it helps denoising algorithms advance to the level where they could be used in practice for automatically denoising digital camera images, without any user interaction.

\section*{Acknowledgements}.
This work was supported in part by DARPA MSEE grant FA 8650-11-1-7149.

\section*{References}

\bibliographystyle{elsarticle-num} 
\bibliography{mybibfile}

\begin{thebibliography}{10}
\expandafter\ifx\csname url\endcsname\relax
  \def\url#1{\texttt{#1}}\fi
\expandafter\ifx\csname urlprefix\endcsname\relax\def\urlprefix{URL }\fi
\expandafter\ifx\csname href\endcsname\relax
  \def\href#1#2{#2} \def\path#1{#1}\fi

\bibitem{dxomark}
D.~Labs, Dxomark sensor scores,
  \url{http://www.dxomark.com/About/Sensor-scores/Use-Case-Scores/}, [Online;
  accessed December 13, 2016] (2009).

\bibitem{buades2005non}
A.~Buades, B.~Coll, J.-M. Morel, A non-local algorithm for image denoising, in:
  CVPR, Vol.~2, 2005, pp. 60--65.

\bibitem{portilla2003image}
J.~Portilla, V.~Strela, M.~J. Wainwright, E.~P. Simoncelli, Image denoising
  using scale mixtures of gaussians in the wavelet domain, Image Processing,
  IEEE Transactions on 12~(11) (2003) 1338--1351.

\bibitem{dabov2007image}
K.~Dabov, A.~Foi, V.~Katkovnik, K.~Egiazarian, Image denoising by sparse 3-d
  transform-domain collaborative filtering, Image Processing, IEEE Transactions
  on 16~(8) (2007) 2080--2095.

\bibitem{barbu2009training}
A.~Barbu, Training an active random field for real-time image denoising, Image
  Processing, IEEE Transactions on 18~(11) (2009) 2451--2462.

\bibitem{StochasticImageDenoising}
F.~Estrada, D.~Fleet, A.~Jepson, Stochastic image denoising (2009).

\bibitem{mairal2009non}
J.~Mairal, F.~Bach, J.~Ponce, G.~Sapiro, A.~Zisserman, Non-local sparse models
  for image restoration, in: ICCV, 2009, pp. 2272--2279.

\bibitem{NeuralNetworkCompete}
H.~C. Burger, C.~J. Schuler, S.~Hamerling, Image denoising: Can plain neural
  networks compete with bm3d?, in: CVPR, 2012, pp. 2392 -- 2399.

\bibitem{schmidt2010generative}
U.~Schmidt, Q.~Gao, S.~Roth, A generative perspective on mrfs in low-level
  vision, in: CVPR, 2010, pp. 1751--1758.

\bibitem{optMRF}
Y.~Chen, T.~Pock, R.~Ranftl, H.~Bischof, Revisiting loss-specific training of
  filter-based mrfs for image restoration, in: German Conference Pattern
  Recognition, 2013, pp. 271--281.

\bibitem{Foi-Poisson}
A.~Foi, M.~Trimeche, V.~Katkovnik, K.~Egiazarian, Practical poissonian-gaussian
  noise modeling and fitting for single-image raw-data, Image Processing, IEEE
  Transactions on 17~(10) (2008) 1737--1754.

\bibitem{NoiseEstimation}
C.~Liu, R.~Szeliski, S.~B. Kang, C.~L. Zitnick, W.~T. Freeman, Automatic
  estimation and removal of noise from a single image 30 (2008) 299--314.

\bibitem{perceptualdistortion}
P.~C. Teo, D.~J. Heeger, Perceptual image distortion, in: International
  Symposium on Electronic Imaging: Science and Technology, 1994, pp. 127--141.

\bibitem{makitalo2014noise}
M.~M{\"a}kitalo, A.~Foi, Noise parameter mismatch in variance stabilization,
  with an application to poisson--gaussian noise estimation, IEEE Transactions
  on Image Processing 23~(12) (2014) 5348--5359.

\bibitem{ponomarenko2015image}
N.~Ponomarenko, L.~Jin, O.~Ieremeiev, V.~Lukin, K.~Egiazarian, J.~Astola,
  B.~Vozel, K.~Chehdi, M.~Carli, F.~Battisti, et~al., Image database tid2013:
  Peculiarities, results and perspectives, Signal Processing: Image
  Communication 30 (2015) 57--77.

\bibitem{ssim}
Z.~Wang, A.~C. Bovik, H.~R. Sheikh, E.~P. Simoncelli, Image quality
  assessment:from error visibility to structural similarity 13 (2004) 600--612.

\bibitem{ImageDenoisingBenchmark}
F.~Estrada, D.~Fleet, A.~Jepson, Image denoising benchmark,
  \url{http://www.cs.utoronto.ca/~strider/Denoise/Benchmark/}, [Online;
  accessed 15-April-2014] (2010).

\bibitem{Luisier2010d}
F.~Luisier, B.~Thierry, M.~Unser, Image denoising in mixed poisson-gaussian
  noise, Image Processing, IEEE Transactions on 20 (2011) 696--708.

\bibitem{ishii2007denoising}
Y.~Ishii, T.~Saito, T.~Komatsu, Denoising via nonlinear image decomposition for
  a digital color camera, in: ICIP, Vol.~1, 2007, pp. I--309.

\bibitem{press2007numerical}
W.~H. Press, Numerical recipes 3rd edition: The art of scientific computing,
  Cambridge University Press, 2007.

\bibitem{RadiometricCCD}
G.~E. Healey, R.~Kondepudy, Radiometric ccd camera calibration and noise
  estimation 16 (1994) 267--276.

\bibitem{vsnr}
D.~M. Chandler, S.~S. Hemami, Vsnr: A wavelet-based visual signal-to-noise
  ratio for natural images 16 (2007) 2284 --2298.

\bibitem{similaritylowlight}
F.~Alter, Y.~Matsushita, X.~Tang, An intensity similarity measure in low-light
  conditions, in: ECCV, 2006, pp. 267--80.

\bibitem{JS-NLM}
Y.~Wu, B.~Tracey, P.~Natarajan, J.~P. Noonan, James--stein type center pixel
  weights for non-local means image denoising, IEEE Signal Processing Letters
  20~(4) (2013) 411--414.

\bibitem{lu2015non}
L.~Lu, W.~Jin, X.~Wang, Non-local means image denoising with a soft threshold,
  IEEE Signal Processing Letters 22~(7) (2015) 833--837.

\bibitem{Shrinkagefields}
U.~Schmidt, S.~Roth, Shrinkage fields for effective image restoration, in:
  CVPR, 2014.

\bibitem{Loss-specific}
J.~Jancsary, S.~Nowozin, C.~Rother, Loss-specific training of nonparametric
  image restoration models: A new state of the art, in: ECCV, 2012, pp.
  112--125.

\end{thebibliography}
\end{document}